\documentclass{article}
\usepackage{float}
\usepackage{fix-cm}
\usepackage{pgf}
\usepackage{iclr2027_conference,times}

\usepackage{amsmath,amsfonts,bm}

\def\eqref#1{equation~\ref{#1}}

\def\1{\bm{1}}

\DeclareMathAlphabet{\mathsfit}{\encodingdefault}{\sfdefault}{m}{sl}
\SetMathAlphabet{\mathsfit}{bold}{\encodingdefault}{\sfdefault}{bx}{n}

\usepackage{booktabs}
\usepackage{multirow}
\usepackage{wrapfig}
\usepackage{capt-of}
\usepackage{hyperref}
\usepackage{url}
\usepackage{graphicx}\usepackage{float}\usepackage{placeins}
\usepackage{capt-of}
\usepackage{amsmath,amssymb}
\usepackage{svg}
\usepackage[ruled,vlined,linesnumbered]{algorithm2e}
\usepackage[table]{xcolor}
\usepackage{relsize}
\definecolor{datasetgray}{RGB}{232,232,232}
\definecolor{heatblue}{RGB}{105,155,205}
\definecolor{heatpink}{RGB}{225,155,175}

\newcommand{\best}[1]{\textbf{#1}}
\newcommand{\second}[1]{\underline{#1}}

\newcommand{\heatcell}[5]{%
  \pgfmathsetmacro{\heatpos}{%
    (#5 == 1) ?
      min(1, max(0, (#1-#2)/max(#3-#2,0.000001))) :
      min(1, max(0, (#3-#1)/max(#3-#2,0.000001)))%
  }%
  \pgfmathparse{\heatpos <= 0.5 ? 1 : 0}%
  \ifnum\pgfmathresult=1
    \pgfmathtruncatemacro{\heatlevel}{%
      round(34*(1-2*\heatpos))%
    }%
    \edef\heatcmd{%
      \noexpand\cellcolor{heatblue!\heatlevel}%
    }%
  \else
    \pgfmathtruncatemacro{\heatlevel}{%
      round(34*(2*\heatpos-1))%
    }%
    \edef\heatcmd{%
      \noexpand\cellcolor{heatpink!\heatlevel}%
    }%
  \fi
  \heatcmd #4%
}

\newcommand{\hval}[4]{%
  \heatcell{#1}{#2}{#3}{#1}{#4}%
}

\newcommand{\hbest}[4]{%
  \heatcell{#1}{#2}{#3}{\best{#1}}{#4}%
}

\newcommand{\hsecond}[4]{%
  \heatcell{#1}{#2}{#3}{\second{#1}}{#4}%
}

\definecolor{apaccent}{RGB}{47,85,151}
\newcommand{\tocheading}[1]{%
  \begingroup
  \color{apaccent}\rule{\linewidth}{0.7pt}\par\vspace{4pt}
  {\centering\normalfont\Large\bfseries\color{black}#1\par}
  \vspace{4pt}\color{apaccent}\rule{\linewidth}{0.7pt}\par
  \endgroup
  \vspace{12pt}
}

\newcommand{\tocdots}{\leaders\hbox to .55em{\hfil\textcolor{gray!55}{.}\hfil}\hfill}
\newcommand{\tocmain}[3]{%
  \noindent
  \textcolor{apaccent}{\textbf{Appendix #1}}\quad
  \hyperref[#3]{\textbf{#2}}\ \tocdots\ \textcolor{gray!65}{\pageref{#3}}\par
  \vspace{4pt}
}

\newcommand{\tocsub}[3]{%
  \hspace{1.8em}\textcolor{apaccent!75!black}{\itshape #1}\quad
  \hyperref[#3]{#2}\ \tocdots\ {\small\textcolor{gray!65}{\pageref{#3}}}\par
  \vspace{2pt}
}

\title{Refinement Is Inherently Editable: Training-Free Prompt-to-Prompt Image Editing with Generative Refinement Network}

\author{%
Yulong Chen$^{\blacktriangle \spadesuit}$ , 
Ziqian Zhang$^{\blacktriangle}$,
Haoyu Zhang$^{\spadesuit}$, 
Ao He$^{\blacktriangle}$, 
Yaxing Wang$^{\diamondsuit}$, 
\textbf{Senmao Li$^{\clubsuit}$}\thanks{Corresponding authors.}~~,
\textbf{Kai Wang$^{\blacktriangle \spadesuit*}$} \\
  \normalfont{$^\blacktriangle$ City University of Hong Kong (Dongguan), Guangdong, China} \\
  \normalfont{$^\spadesuit$ City University of Hong Kong, Hong Kong, China} \\
  \normalfont{$^\clubsuit$ Mohamed bin Zayed University of Artificial Intelligence, Masdar, Abu Dhabi} \\
  \normalfont{$^\diamondsuit$ Jilin University, Changchun, China} \\
{yulong.chen@cityu-dg.edu.cn}\quad
{ziqian.zhang@cityu-dg.edu.cn}\quad 
{hzhang2838-c@my.cityu.edu.hk}\quad \\
{ao.he@cityu-dg.edu.cn}\quad
{yaxing@jlu.edu.cn}\quad
{senmaonk@foxmail.com}\quad
{kai.wang@cityu-dg.edu.cn}\quad
}

\iclrfinalcopy 
\begin{document}

\maketitle
\fancyhead{}
\lhead{Preprint}
\vspace{-0.45cm}
\begin{figure}[htbp]
\centering
\includegraphics[width=0.98\linewidth]{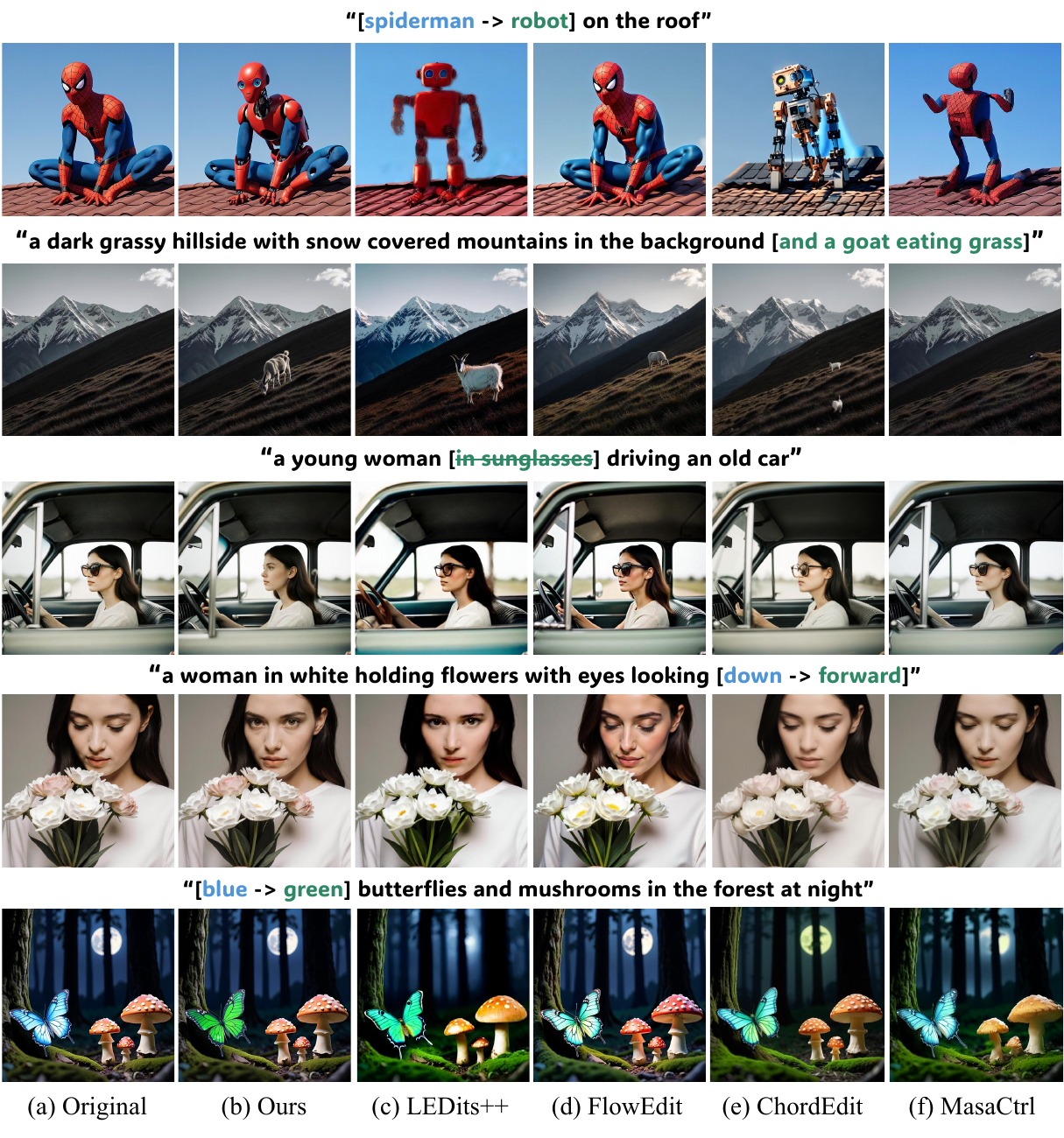}
\vspace{-0.5cm}
\caption{Visual comparisons of text-guided image editing results from our method \textbf{\textit{RefineEdit}}, LEDits++~\citep{leditspp}, FlowEdit~\citep{flowedit}, ChordEdit~\citep{chordedit}, and MasaCtrl~\citep{masactrl}. \textcolor[HTML]{4E95D9}{Blue} and \textcolor[HTML]{2D8864}{green} indicate the original and editing prompts, respectively.
}
\label{fig:preview}

\end{figure}

\begin{abstract}

Text-guided image editing must introduce the requested changes while
preserving unrelated source content. In training-free editing, diffusion editors often use spatial controls whose inaccuracies can leave edits incomplete or alter unrelated regions. Causal autoregressive editors face a further constraint:
their fixed decoding order limits revision of earlier decisions. As the first to explore training-free image editing with Generative Refinement Network (GRN), we observe that its refinement process is inherently suitable for editing and offers a promising way to address these limitations.
Motivated by this observation, we introduce
\textbf{\textit{RefineEdit}}, a training-free prompt-to-prompt image editing framework built on the GRN.
Our key idea is to couple edit localization with content generation
through the global refinement of binary image codes, allowing editing evidence to be revised as the image evolves. 
More specifically, \textbf{\textit{RefineEdit}} combines \textit{bit routing} with two stabilization mechanisms: \textit{adaptive spatial freezing} and \textit{finite bit locking}. \textit{Bit routing} starts from an intermediate source state and uses signed probability differences between the two branches to identify editable positions and bits. It directs selected bits toward editing refinement while anchoring the rest to the evolving source trajectory. \textit{Adaptive spatial freezing} limits unnecessary expansion of the editing region, while \textit{finite bit locking} maintains recent bit activations to support continued editing.
The overall framework requires no additional training, external masks, or attention control. Across nine editing categories of PIE-Bench, \textbf{\textit{RefineEdit}} achieves the best background-preservation scores in PSNR, LPIPS, MSE, and SSIM, together with the highest whole-image and edited-region CLIP scores among the evaluated methods. Code is available at https://github.com/mura1n/RefineEdit.

\end{abstract}

\section{Introduction}
\label{sec:introduction}

Text-to-image models can now generate high-quality images that closely follow text prompts~\citep{latent_diffusion_model, labs2025flux, LlamaGen, stable_diffusion3, infinity}. Yet visual creation is often iterative, and users may revise a generated image rather than start over. Text-guided image editing aims to make the changes specified by an editing prompt while preserving unrelated source content~\citep{chordedit, masactrl, p2p, flowedit, leditspp}. Balancing these goals is difficult: stronger edits may alter unrelated regions, while stronger preservation may weaken the intended change. The key challenge is to identify where the source conflicts with the editing prompt and make the required change while preserving other content.

Most training-free editors~\citep{Avrahami_2025_CVPR, Fu_2025_CVPR} address this challenge by adding spatial control to a pretrained generator. For example, diffusion methods use masks, attention maps, or injected features~\citep{p2p,blended,masactrl,instructpix2pix,pnp}, while autoregressive (AR) methods select between source and editing predictions during decoding~\citep{aredit}. These controls determine which parts of the source may change, making their accuracy and evolution important to the editing quality: an overly narrow editing region may leave the requested change incomplete, whereas an overly broad region may alter unrelated content. 
Causal AR models make generation or editing decisions in early stages, which are difficult to revise~\citep{esser2021taming,LlamaGen,kondratyuk2023videopoet,var,wang2026editinfinity}. 
This raises a question: \textbf{\textit{Can the evolving generation process itself provide the necessary evidence to revise the edit localization?}}

\begin{figure}[t!]
\centering
\includegraphics[width=\linewidth]{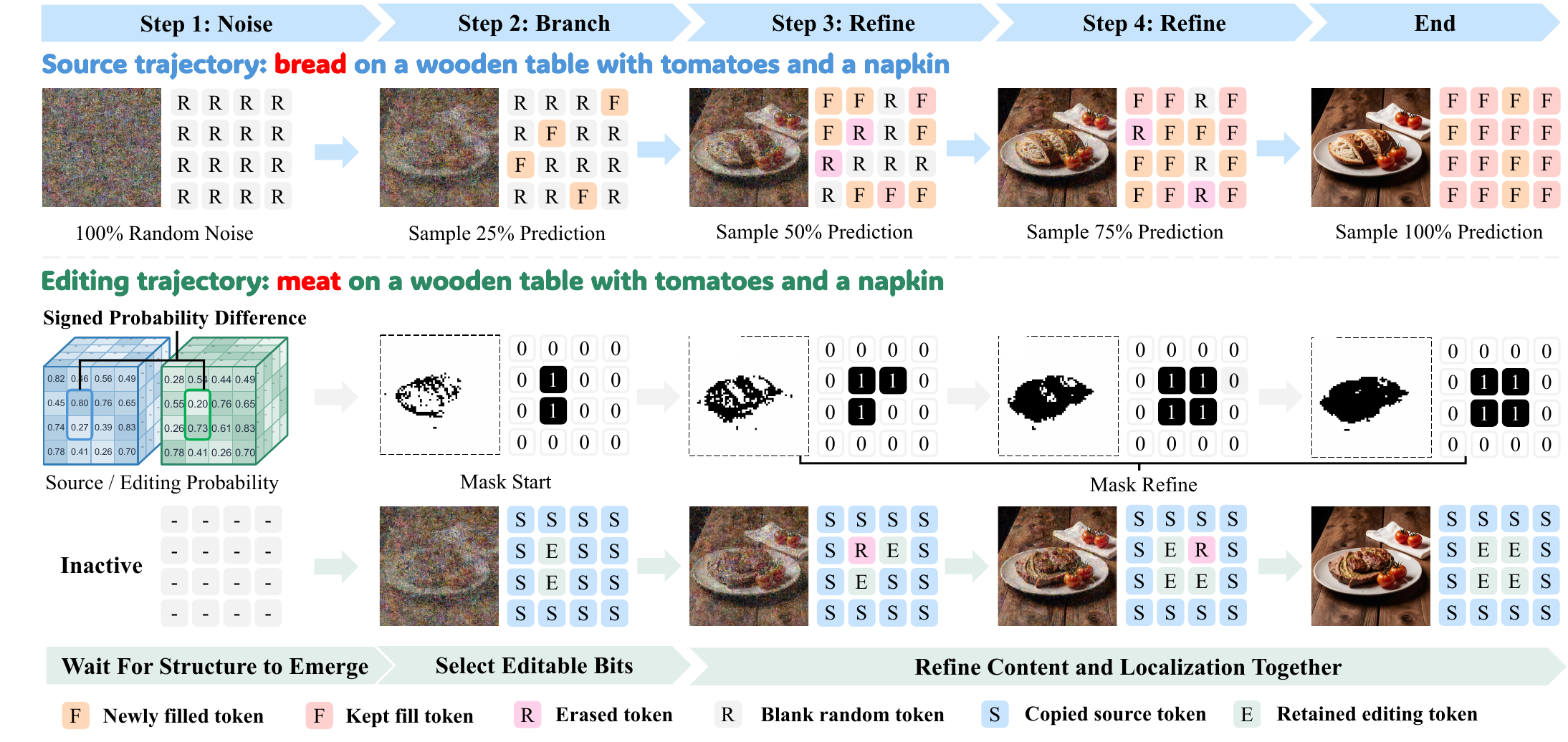}
\vspace{-0.65cm}
\caption{
\textbf{\textit{Top:}} GRN progressively develops image structure from a
random binary code.
\textbf{\textit{Middle:}} Source and editing probabilities, illustrated
by the blue and green cubes, are compared at the same
sampled bit values.
Black mask entries indicate editable positions.
\textbf{\textit{Bottom:}} Selected bits follow editing refinement,
while unselected bits copy the evolving source state.}
\label{fig:moti}
\vspace{-0.55cm}
\end{figure}

The Generative Refinement Network (GRN)~\citep{grn} offers a viable setting for exploring this problem. As shown in Fig.~\ref{fig:moti}, GRN repeatedly updates the full visual token map, moving from noise\footnote{Noise refers
to a random binary code whose bits are independently sampled
as 0 or 1 with equal probability.}
to coarse structure and then fine details. Intermediate source states already contain useful spatial structure, allowing the editing branch to reuse this layout while revising object appearance. This resembles an artist developing the same rough feline sketch into a cat or a tiger. 
Beyond providing a reusable layout, GRN represents images with Hierarchical Binary Quantization (HBQ) codes whose spatial and channel coordinates align across branches. 
At the branching step, conditioning the shared state on the source and editing prompts yields different bit probabilities. 
Such differences can serve as the editing evidence at individual binary coordinates, suggesting a way to select editing updates while retaining source bits elsewhere. 
As refinement changes the edited state, subsequent predictions provide new evidence for bit selection. These selections guide the next content update, allowing edit
localization and content generation to evolve together rather
than remain separate processes.
This leads to our central insight: \textbf{\textit{Refinement is inherently editable.}} These properties make GRN a natural basis for editing through selective bit updates.

Motivated by the above observations, we propose \textbf{\textit{RefineEdit}},
a training-free framework for prompt-to-prompt image editing framework based on GRN.
To the best of our knowledge, \textbf{\textit{RefineEdit}} is the first method to
adapt GRN to training-free image editing, 
extending its original applications in
image generation and training-based video
editing~\citep{grnedit}. 
Specifically, at the switch step of \textbf{\textit{RefineEdit}}, it initializes an editing branch from the current source state. Both the source and editing branches continue refinement with their corresponding prompts. 
From the switch step onward, we compare the probabilities assigned by the two branches to the same bits sampled from the source branch (source-sampled bits) at each refinement step. We use the resulting signed probability differences to construct a spatial mask that selects editable positions and a bitwise mask that determines which bits at those positions may change.
\textit{Source-anchored bit routing} accepts editing updates at selected coordinates and copies the evolving source state at all remaining coordinates.
To stabilize these routing decisions across refinement steps, we therefore introduce \textit{adaptive spatial freezing (AdaSF)},
which freezes spatial selection at the switch step when the
initial editing response is sufficiently strong.
Meanwhile, \textit{finite bit locking (FBL)} retains editing
permission for recently selected bits over a short window,
so temporary probability fluctuations do not immediately
interrupt their refinement. These two novel techniques in our method \textbf{\textit{RefineEdit}} operate directly on binary codes, without external editing masks or attention control. 

In our experiments, we compare \textbf{\textit{RefineEdit}} with existing methods across nine editing categories of PIE-Bench~\citep{directinversionandpiebench}. \textbf{\textit{RefineEdit}} achieves the best background-preservation scores in PSNR, LPIPS, MSE, and SSIM, together with the highest whole-image and edited-region CLIP scores among the compared methods.
To summarize, our contributions are as follows:
\begin{itemize}
    \item Based on our observation that refinement is inherently editable, we introduce \textbf{\textit{RefineEdit}}, the first training-free prompt-to-prompt image editing framework built on GRN. RefineEdit couples edit localization with content generation, allowing editing evidence to evolve with the visual state.
    \item We develop source-anchored bit routing, which uses signed probability differences to select editable positions and bits while anchoring the remaining bits to the evolving source. \textit{AdaSF} limits unnecessary expansion of the editing region, and \textit{FBL} maintains  editing permission for recently selected bits to support continued refinement.
    \item Experiments across nine editing categories of PIE-Bench demonstrate that \textbf{\textit{RefineEdit}} effectively follows editing prompts while preserving unrelated source content. Ablation studies further show the complementary roles of AdaSF and FBL in stabilizing edit localization and supporting bitwise editing.
\end{itemize}

\section{Related Work}
\label{relate}
\subsection{Text-to-Image Generation}

Text-to-image generation has evolved through several generative paradigms. Diffusion and flow models have established strong image fidelity and text alignment~\citep{dit,SDTurbo,sdxl,latent_diffusion_model,stable_diffusion3,sit}. Recent autoregressive models show that, with improved visual tokenizers and model scaling, discrete next-token prediction can achieve competitive or even superior generation quality~\citep{zero_shot_DALLE,Parti,LlamaGen}. However, their causal order prevents earlier tokens from being revised once generated. Visual autoregressive models reorganize generation as next-scale or bitwise prediction~\citep{var,infinity}, improving efficiency and discrete modeling but still committing to a fixed coarse-to-fine order. GRN~\citep{grn} overcomes this restriction through near-lossless Hierarchical Binary Quantization and global random refinement, allowing every binary coordinate to be repeatedly reconsidered. Its globally revisable binary representation provides a natural interface for image editing, enabling changes to be localized and controlled at the bit level.
\subsection{Training-free image editing}
Training-free image editing repurposes pretrained generators without parameter updates. Diffusion approaches commonly use mask blending, attention control, or intermediate-feature injection to introduce target semantics while retaining source structure~\citep{p2p,blended,masactrl,instructpix2pix,pnp,li2023stylediffusion,leditspp,chordedit,WU2026114063}. However, such continuous control often trades edit strength for background fidelity. Flow methods improve semantic transport and efficiency by constructing paths between source and target distributions~\citep{flowedit,rfinversion}, yet editable support remains implicit rather than enforced at native coordinates. Discrete autoregressive editing instead exploits token-distribution differences to localize changes~\citep{aredit,atm}, but its control follows a fixed causal order and does not directly match GRN's globally revisable generation~\citep{grnedit}. 
In this paper, our proposed \textbf{\textit{RefineEdit}} uses aligned-trajectory contrast to route only prompt-sensitive bits to editing and anchor the rest to the evolving source. 

\section{Preliminaries}

\paragraph{\textit{Binary image representation.}}
A frozen Hierarchical Binary Quantization (HBQ) tokenizer $\mathcal{E}_{\mathrm{HBQ}}$ encodes an image $I$ as $Y=\mathcal{E}_{\mathrm{HBQ}}(I)\in\{0,1\}^{N\times D}$, where $N$ is the number of spatial positions and $D$ is the number of bits per position.
Each coordinate $(n,d)$ identifies one bit at spatial position $n$.

\paragraph{\textit{Generative refinement.}}
GRN initializes $Y_0=Z$, where each bit of $Z$ is sampled once as $0$ or $1$ with equal probability. The same random code is reused throughout refinement.
Given the current binary input \(Y_t\) and text condition \(c\), GRN predicts all binary coordinates in parallel:
\begin{equation}
    P_t
    =
    \operatorname{softmax}\!\left(G_\theta(Y_t,c,t)\right),
    \qquad
    \widehat{Y}_t\sim\operatorname{Cat}(P_t).
    \label{eq:grn_prediction}
\end{equation}
Here, \(G_\theta\) is the frozen GRN Transformer, which outputs two logits per binary coordinate. Softmax converts these logits into the distribution \(P_t(n,d,\cdot)\) over \(\{0,1\}\). \(\operatorname{Cat}\) samples each bit of \(\widehat Y_t\) from its corresponding distribution.

The next state combines sampled predictions with the fixed random code:
\begin{equation}
    Y_{t+1}
    =
    S_t\odot\widehat{Y}_t
    +(1-S_t)\odot Z,
    \label{eq:grn_refinement}
\end{equation}
where $\odot$ denotes the element-wise multiplication operator.
The selection mask \(S_t\) satisfies
$S_{t,n,d}\sim\operatorname{Bernoulli}(\lambda_{t+1})$:
it equals $1$ with probability $\lambda_{t+1}$ and $0$ otherwise.
Thus, $\lambda_{t+1}$ controls the expected proportion of
predictions retained in the next state.
As this proportion increases, refinement moves from random bits
toward a complete image. The selection mask \(S_t\) is resampled at every step, so a coordinate may return to its initial random value. All coordinates are predicted again at the next step rather than permanently fixed.

\section{Method}
\label{method}

\begin{figure}[t]
    \centering
    \includegraphics[width=\linewidth]{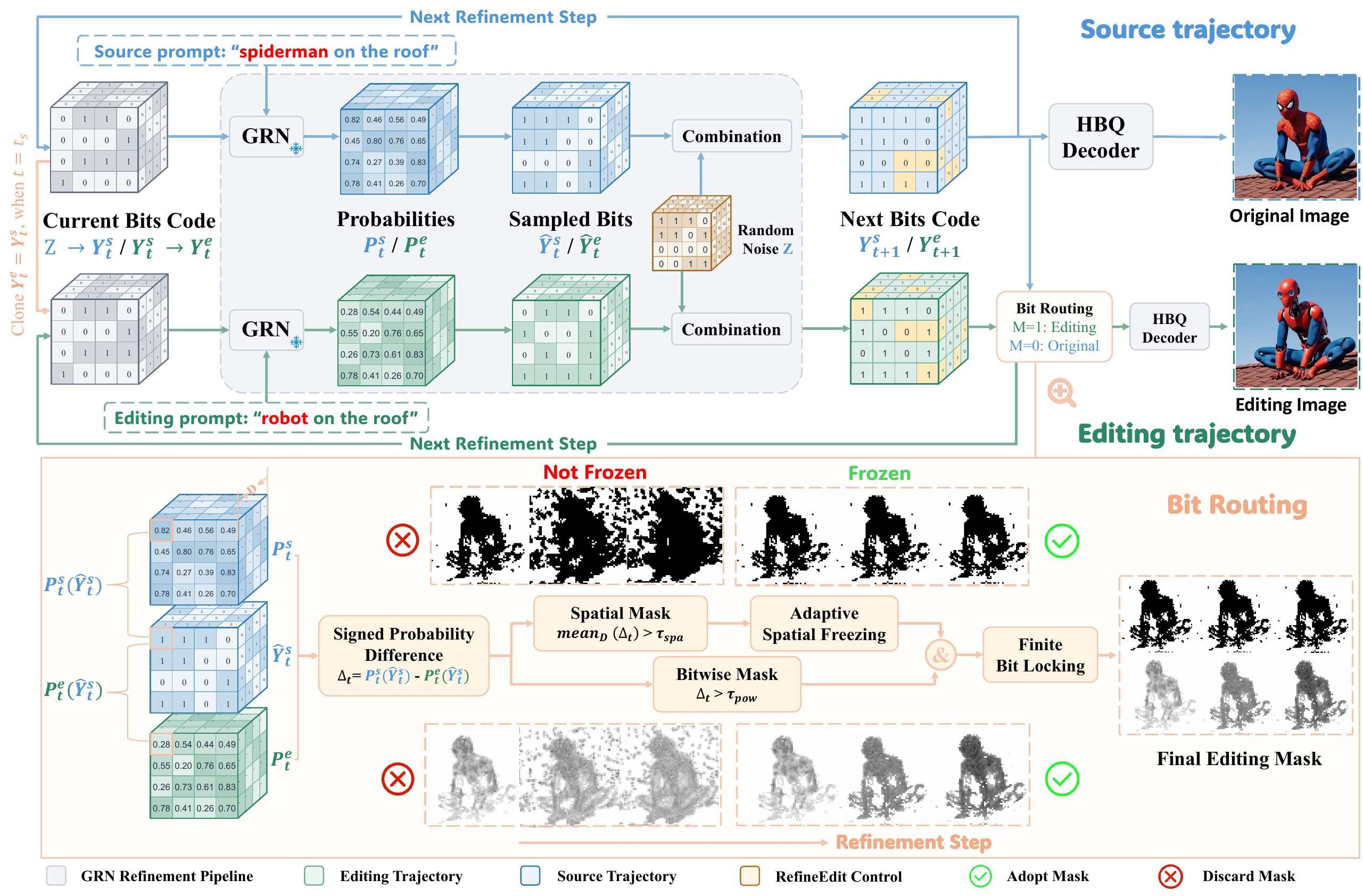}
    \vspace{-0.6cm}
    \caption{
    Overview of \textbf{\textit{RefineEdit}}.
    The editing branch inherits the source state at the switch step.
    Both branches retain GRN's global prediction and random refinement,
    while bit routing selects editing updates or copies the evolving
    source state.
    }
    \label{fig:framework}
    \vspace{-0.3cm}
\end{figure}

In this section, we present \textbf{\textit{RefineEdit}}, a training-free
prompt-to-prompt image editing framework built on GRN.
Section~\ref{sec:bit_routing} describes how we derive editing
masks from aligned bit probabilities and use them to route
source and editing updates.
Section~\ref{sec:stabilizing_routing} introduces spatial and
temporal stabilization to limit mask expansion and sustain
bitwise editing.
Figure~\ref{fig:framework} provides an overview.

\subsection{Refinement-Guided Bit Routing}
\label{sec:bit_routing}

Given a source prompt $c^s$ and an editing prompt $c^e$,
RefineEdit first runs source refinement under $c^s$.
At the switch step $t_s$, we initialize the editing branch
from the current source state,
$Y_{t_s}^e=Y_{t_s}^s$.
Both branches then continue refinement under their corresponding
prompts, using the same fixed random code $Z$ and refinement
schedule.
The editing branch therefore inherits the intermediate structure from the source branch. We use differences between the two branches' bit probabilities as editing evidence to identify coordinates that may depart from the source.

\paragraph{\textit{Editing evidence.}}
At each editing step, the two branches produce \(P_t^s\) and \(P_t^e\) with Eq.~\ref{eq:grn_prediction}.
We denote the source-sampled bit at coordinate $(n,d)$ as
$\widehat{y}_{t,n,d}=\widehat{Y}_{t,n,d}^s$.
Its signed probability difference is:
\begin{equation}
    \Delta_{t,n,d}
    =
    P_t^s(n,d,\widehat{y}_{t,n,d})
    -
    P_t^e(n,d,\widehat{y}_{t,n,d}).
    \label{eq:signed_probability_drop}
\end{equation}
A positive value means that the editing branch assigns less
probability to the source-sampled bit, providing evidence
for allowing that coordinate to depart from the source.
The branches share the same input only at the switch step. 
During the following refinement steps, the probability differences reflect both the respective prompts and the evolving branch states. We use these differences to determine where editing is allowed and which bits may change.

\paragraph{\textit{Spatial and bitwise mask selection.}}
We express the two decisions described above through
instantaneous spatial and bitwise masks. Averaging the evidence across all $D$ bits at each position gives the spatial score $q_{t,n}=D^{-1}\sum_{d=1}^{D}\Delta_{t,n,d}$.
The spatial mask $a_t\in\{0,1\}^{N}$ and bitwise mask
$m_t\in\{0,1\}^{N\times D}$ are then defined as:
\begin{equation}
    \begin{aligned}
        a_{t,n}
        =
        \mathbf{1}[q_{t,n}>\tau_{\mathrm{spa}}],\qquad
        m_{t,n,d}
        =
        a_{t,n}\,
        \mathbf{1}[\Delta_{t,n,d}>\tau_{\mathrm{pow}}],
    \end{aligned}
    \label{eq:spatial_bit_masks}
\end{equation}
where $\mathbf{1}[\cdot]$ equals $1$ when its condition holds
and $0$ otherwise.
The spatial threshold selects positions, while the power threshold determines which bits within those positions receive editing permission.
For a fixed spatial mask and fixed probabilities, lowering \(\tau_{\mathrm{pow}}\) relaxes bit selection. Changing these selections alters subsequent branch states and predictions, so bitwise control can also influence later spatial masks. These masks specify editing permissions, which we apply through source-anchored routing.

\paragraph{\textit{Source-anchored routing.}}
Standard GRN refinement in Eq.~\ref{eq:grn_refinement} produces the next source state \(Y_{t+1}^s\) and an editing proposal \(\widetilde Y_{t+1}^e\). We then route their bits using \(m_t\):
\begin{equation}
    Y_{t+1}^e
    =
    m_t\odot\widetilde{Y}_{t+1}^e
    +(1-m_t)\odot Y_{t+1}^s.
    \label{eq:source_anchored_routing}
\end{equation}
Selected coordinates take values from the editing proposal; all other coordinates copy the source state at the same refinement step. Both branches therefore retain global GRN refinement; the mask controls which updates enter the next editing state. At the final step, we route the sampled source and editing predictions and decode the resulting HBQ code, without mixing in \(Z\).

Eq.~\ref{eq:spatial_bit_masks} selects masks from the current predictions. As the predictions evolve, the spatial mask may expand into unrelated regions, and previously selected bits may lose editing permission. Section~\ref{sec:stabilizing_routing} addresses these issues by two novel techniques before the masks are used in routing.

\subsection{Mask stabilization}
\label{sec:stabilizing_routing}

\paragraph{\textit{Adaptive spatial freezing (AdaSF).}}
As mentioned above, mask expansion can introduce changes in unrelated regions. To address this problem, we adapt the freezing decision to each sample using its
initial editing response. This response determines whether to freeze the spatial mask at the switch step or continue updating it. It is formulated as:
\begin{equation}
    \Omega=\{n\mid a_{t_s,n}=1\},
    \qquad
    r=
    \frac{
        \sum_{n\in\Omega}
        (q_{t_s,n}-\tau_{\mathrm{spa}})
    }{
        \max(1,|\Omega|)
    },
    \label{eq:initial_response}
\end{equation}
Here, \(\Omega\) contains the positions selected at the switch step, and \(|\Omega|\) is their number. The response \(r\) measures their average score margin above \(\tau_{\mathrm{spa}}\), with \(r=0\) when no position is selected.
When $r\geq\tau_{\mathrm{frz}}$, we retain the initial
spatial mask by setting $a_t=a_{t_s}$ for all $t\geq t_s$.
Otherwise, $a_t$ continues to follow
Eq.~\ref{eq:spatial_bit_masks}.
We use $\tau_{\mathrm{f r z}}=2\tau_{\mathrm{spa}}$
by default.

The decision is made once at the switch step and affects only spatial selection; bitwise selection and content refinement continue. Even within a frozen spatial mask, changes in bit
probabilities can deactivate previously selected bits.

\paragraph{\textit{Finite bit locking (FBL).}}
To reduce these interruptions, FBL retains a bit's editing permission if it was selected by the instantaneous bitwise
mask in Eq.~\ref{eq:spatial_bit_masks} at least once within the latest $K$ steps. This replaces the instantaneous rule for $m_t$ with: 
\begin{equation}
    m_{t,n,d}
    =
    \max_{\max(t_s,t-K+1)\leq j\leq t}
    \left\{
        a_{j,n}\,
        \mathbf{1}[\Delta_{j,n,d}>\tau_{\mathrm{pow}}]
    \right\}.
    \label{eq:finite_bit_lock}
\end{equation}
Here, $j$ indexes the latest finite $K$ editing steps, including
the current step.
The maximum implements a logical OR over their instantaneous
activations.
Setting $K=1$ recovers the instantaneous rule. When spatial selection remains adaptive, a recently activated
bit may remain editable even if its position is excluded by
the current spatial mask.
Thus, the spatial mask gates instantaneous activations,
while $m_t$ also retains recent ones.

Locking preserves editing permission, not the bit value.
Selected bits therefore continue to undergo prediction and
random refinement.
A bit loses permission once it has not passed the selection
tests for $K$ consecutive steps.

Overall, bit routing, AdaSF, and FBL balance source
preservation with continued refinement of edited content.
Bit routing anchors unselected bits to the evolving source,
while AdaSF and FBL address unnecessary mask expansion and
interruptions to bitwise editing.
The complete training-free editing procedure is given in
Algorithm~\ref{alg:refineedit}.


\section{Experiment}
\label{exp}
\subsection{Experimental Setups}
\paragraph{\textit{Benchmarks.}}
\textbf{\textit{RefineEdit}} uses a pretrained GRN without additional training
and generates images at $1024 \times 1024$ resolution.
We evaluate it using the source and editing prompt pairs from
nine editing categories of PIE-Bench~\citep{directinversionandpiebench}: object replacement,
addition and removal, content and pose modification, color and
material modification, background modification, and style transfer.
All prompts are taken directly from PIE-Bench.
For each prompt pair, GRN generates a source image from the source prompt.
All baselines edit this same source image using the corresponding editing
prompt, providing a common source reference across methods. We use Grounded-SAM~\citep{ren2024grounded} to obtain foreground masks for regional
evaluation. These masks are shared across methods and are separate
from the dynamic masks predicted by \textbf{\textit{RefineEdit}}.
Since the baselines produce $512 \times 512$ images in our setup,
we resize the GRN source images and \textbf{\textit{RefineEdit}} outputs to
$512 \times 512$ before evaluation.
The evaluation masks are aligned to the same resolution. We additionally evaluate on EditEval v2~\citep{imageEditingSurvey} using the
same settings from PIE-Bench without retuning.
Details of the PIE-Bench and EditEval v2 evaluations are provided
in Appendices~\ref{app:dataset} and~\ref{app:editeval}, respectively.

\paragraph{\textit{Evaluation Metrics.}}
We evaluate structural consistency, preservation of unedited
content, and alignment with the editing prompt.
Structure Distance~\citep{tumanyan2022splicing} measures structural
differences between source and edited images.
PSNR~\citep{wang2004image}, LPIPS~\citep{zhang2018unreasonable}, MSE~\citep{wang2009mean}, and SSIM~\citep{wang2004image} measure content
preservation in unedited regions.
We assess semantic alignment using CLIP~\citep{radford2021learning} similarity
between the editing prompt and the edited image, computed over
both the whole image and the designated editing region.
We denote these two scores by $\mathrm{CLIP}_{\mathrm{tgt}}$
and $\mathrm{CLIP}_{\mathrm{edit}}$, respectively.
All methods use the same evaluation images, masks, and metric
implementations. Metrics details are provided in Appendix~\ref{app:metrics}.

\vspace{-0.3cm}

\begin{table*}[htbp]
\centering
\caption{Quantitative comparison across nine editing categories of PIE-Bench. Methods are grouped by generation paradigm. \textbf{Bold} and \underline{underlined} values indicate the best and second-best results, respectively. Background colors represent a diverging heatmap normalized by \textit{performance}: \textcolor{heatblue}{Blue} for worst, white for average, and \textcolor{heatpink}{Pink} for best.}
\label{tab:quantitative_comparison}

\setlength{\tabcolsep}{6pt}
\fontsize{7.4}{8.4}\selectfont

\resizebox{\linewidth}{!}{%
\begin{tabular}{ll c cccc cc}
\toprule
\multirow{2}{*}{\textbf{Method}} & \multirow{2}{*}{\textbf{Model}} & \multicolumn{1}{c}{\textbf{Struct.}} & \multicolumn{4}{c}{\textbf{Background Preservation}} & \multicolumn{2}{c}{\textbf{CLIP Sim.}} \\[-0.15em]
& & \textbf{Dist.}$\downarrow$ & \textbf{PSNR}$\uparrow$ & \textbf{LPIPS}$\downarrow$ & \textbf{MSE}$\downarrow$ & \textbf{SSIM}$\uparrow$ & \textbf{Whole}$\uparrow$ & \textbf{Edited}$\uparrow$ \\
\cmidrule(lr){3-3} \cmidrule(lr){4-7} \cmidrule(lr){8-9}
\midrule

\rowcolor{datasetgray}
\multicolumn{9}{@{}l}{\hspace{0.25em}\textbf{\textit{Diffusion Methods}}} \\[-0.1em]

P2P~\citep{p2p} & SD 1.4 & \hval{0.0572}{0.0162}{0.0798}{0} & \hval{19.02}{19.02}{30.36}{1} & \hval{0.149}{0.037}{0.176}{0} & \hval{0.0172}{0.0046}{0.0184}{0} & \hval{0.771}{0.711}{0.950}{1} & \hval{24.88}{23.94}{26.59}{1} & \hval{22.26}{21.78}{23.44}{1} \\
MasaCtrl~\citep{masactrl} & SD 1.4 & \hval{0.0297}{0.0162}{0.0798}{0} & \hval{22.25}{19.02}{30.36}{1} & \hval{0.092}{0.037}{0.176}{0} & \hval{0.0088}{0.0046}{0.0184}{0} & \hval{0.838}{0.711}{0.950}{1} & \hval{24.30}{23.94}{26.59}{1} & \hval{21.78}{21.78}{23.44}{1} \\
Pix2Pix-Zero~\citep{parmar2023zero} & SD 1.4 & \hval{0.0798}{0.0162}{0.0798}{0} & \hval{19.45}{19.02}{30.36}{1} & \hval{0.170}{0.037}{0.176}{0} & \hval{0.0184}{0.0046}{0.0184}{0} & \hval{0.771}{0.711}{0.950}{1} & \hval{23.94}{23.94}{26.59}{1} & \hval{21.92}{21.78}{23.44}{1} \\
PnP~\citep{pnp} & SD 1.5 & \hval{0.0271}{0.0162}{0.0798}{0} & \hval{22.80}{19.02}{30.36}{1} & \hval{0.091}{0.037}{0.176}{0} & \hval{0.0074}{0.0046}{0.0184}{0} & \hval{0.842}{0.711}{0.950}{1} & \hval{25.51}{23.94}{26.59}{1} & \hval{22.74}{21.78}{23.44}{1} \\
PnP-DirectInv~\citep{directinversionandpiebench} & SD 1.5 & \hsecond{0.0229}{0.0162}{0.0798}{0} & \hval{23.37}{19.02}{30.36}{1} & \hval{0.081}{0.037}{0.176}{0} & \hval{0.0064}{0.0046}{0.0184}{0} & \hval{0.852}{0.711}{0.950}{1} & \hval{25.57}{23.94}{26.59}{1} & \hval{22.77}{21.78}{23.44}{1} \\
LEDits++~\citep{leditspp} & SD 1.5 & \hval{0.0414}{0.0162}{0.0798}{0} & \hval{20.59}{19.02}{30.36}{1} & \hval{0.100}{0.037}{0.176}{0} & \hval{0.0120}{0.0046}{0.0184}{0} & \hval{0.820}{0.711}{0.950}{1} & \hsecond{26.22}{23.94}{26.59}{1} & \hsecond{23.29}{21.78}{23.44}{1} \\
ChordEdit~\citep{chordedit} & SD-Turbo & \hval{0.0266}{0.0162}{0.0798}{0} & \hval{23.02}{19.02}{30.36}{1} & \hval{0.097}{0.037}{0.176}{0} & \hval{0.0071}{0.0046}{0.0184}{0} & \hval{0.812}{0.711}{0.950}{1} & \hval{24.94}{23.94}{26.59}{1} & \hval{22.41}{21.78}{23.44}{1} \\
\addlinespace[0.12em]
\cmidrule(lr){1-9}

\rowcolor{datasetgray}
\multicolumn{9}{@{}l}{\hspace{0.25em}\textbf{\textit{Flow Methods}}} \\[-0.1em]

FlowEdit~\citep{flowedit} & FLUX.1-dev & \hbest{0.0162}{0.0162}{0.0798}{0} & \hsecond{25.03}{19.02}{30.36}{1} & \hsecond{0.062}{0.037}{0.176}{0} & \hsecond{0.0046}{0.0046}{0.0184}{0} & \hsecond{0.902}{0.711}{0.950}{1} & \hval{24.85}{23.94}{26.59}{1} & \hval{21.95}{21.78}{23.44}{1} \\
RF-Inversion~\citep{rfinversion} & FLUX.1-dev & \hval{0.0484}{0.0162}{0.0798}{0} & \hval{19.73}{19.02}{30.36}{1} & \hval{0.176}{0.037}{0.176}{0} & \hval{0.0148}{0.0046}{0.0184}{0} & \hval{0.711}{0.711}{0.950}{1} & \hval{25.04}{23.94}{26.59}{1} & \hval{22.44}{21.78}{23.44}{1} \\
\addlinespace[0.12em]
\cmidrule(lr){1-9}

\rowcolor{datasetgray}
\multicolumn{9}{@{}l}{\hspace{0.25em}\textbf{\textit{GRN Method}}} \\[-0.1em]

\textbf{\textit{RefineEdit (Ours)}} & GRN & \hval{0.0266}{0.0162}{0.0798}{0} & \hbest{30.36}{19.02}{30.36}{1} & \hbest{0.032}{0.037}{0.176}{0} & \hbest{0.0045}{0.0045}{0.0184}{0} & \hbest{0.950}{0.711}{0.950}{1} & \hbest{26.59}{23.94}{26.59}{1} & \hbest{23.44}{21.78}{23.44}{1} \\
\bottomrule
\end{tabular}%
}
\vspace{-0.4cm}
\end{table*}

\paragraph{\textit{Comparison Methods.}}
We compare \textbf{\textit{RefineEdit}} with nine training-free image editing
baselines, grouped by the generative backbones used in our
evaluation. Flow baselines include FlowEdit~\citep{flowedit} and
RF-Inversion~\citep{rfinversion}. Diffusion baselines include MasaCtrl~\citep{masactrl},
Prompt-to-Prompt (P2P)~\citep{p2p}, Pix2Pix-Zero~\cite{parmar2023zero}, PnP~\citep{pnp}, PnP-DirInv~\citep{directinversionandpiebench}, LEDits++~\citep{leditspp}, and ChordEdit~\citep{chordedit}.
Comparison methods details are provided in Appendix~\ref{app:baselines}.

\textbf{\textit{Implementation details.}}
\textbf{\textit{RefineEdit}} uses a frozen pretrained GRN without attention control.
We select $t_s$, $\tau_{\mathrm{spa}}$, and
$\tau_{\mathrm{pow}}$ by editing category, while fixing
$K=4$ and $\tau_{\mathrm{frz}}=2\tau_{\mathrm{spa}}$.
Each inference run uses a single NVIDIA A100 GPU.
Category-specific settings are
provided in Appendix~\ref{app:refineEdit_setting}.

\subsection{Experimental Results}

Figure~\ref{fig:preview} presents a \textit{qualitatively} comparison of
representative editing results.
\textbf{\textit{RefineEdit}} replaces spiderman with a robot while retaining
the crouching pose and roof layout.
LEDits++ and ChordEdit alter the pose or roof, whereas FlowEdit
largely retains the source appearance.
\textbf{\textit{RefineEdit}} also adds a goat while preserving the mountain scene,
removes sunglasses while retaining the car interior, and
changes a woman's gaze without disturbing the bouquet.
For color editing, it turns the butterfly green while retaining
the surrounding composition.
These examples illustrate localized changes with limited
disruption to unrelated content. More \textit{qualitatively} results are provided in Appendix~\ref{app:qualitative_results}.

Table~\ref{tab:quantitative_comparison} reports \textit{quantitative comparison} results over all nine
editing categories. \textbf{\textit{RefineEdit}} achieves the best scores on all four
background-preservation metrics and both CLIP metrics.
Compared with FlowEdit, it improves PSNR by $5.33$\,dB.
Its edited-region CLIP score exceeds LEDits++ by $0.15$,
while whole-image alignment also ranks first.
The gains in MSE and CLIP are modest, and FlowEdit remains
stronger in Structure Distance.
Overall, these results show strong preservation of unedited
content without sacrificing alignment with editing prompts. On EditEval v2, \textbf{\textit{RefineEdit}} also achieves
strong content preservation and competitive alignment with
editing prompts (Appendix~\ref{app:editeval}).

\subsection{Ablations and analysis}

\begin{wraptable}{r}{0.6\linewidth}
    \centering
    \small
    \vspace{-0.8cm}
    \caption{Runtime comparison with two diffusion and two flow editing methods. Type denotes the generation paradigm. Time is the average runtime per editing operation over ten runs, excluding I/O time.}
    \label{tab:efficiency}
    \begin{tabular}{@{}llcr@{}}
        \toprule
        Method & Type & Resolution & Time (s)$\downarrow$ \\
        \midrule
        PnP & Diffusion & $1K$ & 79.70 \\
        PnP-DirectInv & Diffusion & $1K$ & 79.65 \\
        \midrule
        RF-Inversion & Flow & $1K$ & 32.05 \\
        FlowEdit & Flow & $1K$ & 27.15 \\
        \midrule
        RefineEdit (Ours) & GRN & $1K$ & \textbf{26.77} \\
        \bottomrule
    \end{tabular}
    \vspace{-2.0em}
\end{wraptable}

\paragraph{\textit{Latency.}}
We measure runtime on a single NVIDIA A100 GPU.
Each editing operation is repeated ten times, and we report
the average runtime excluding I/O time. 
As shown in Table~\ref{tab:efficiency}, \textbf{\textit{RefineEdit}} takes 26.77 seconds per edit at $1024 \times 1024$ resolution.
Its runtime is comparable to FlowEdit and lower than
RF-Inversion, while providing approximately threefold
speedup over PnP and PnP-DirectInv.

\begin{figure}[t!]
\centering
\vspace{-0.6cm}
\includegraphics[width=\linewidth]{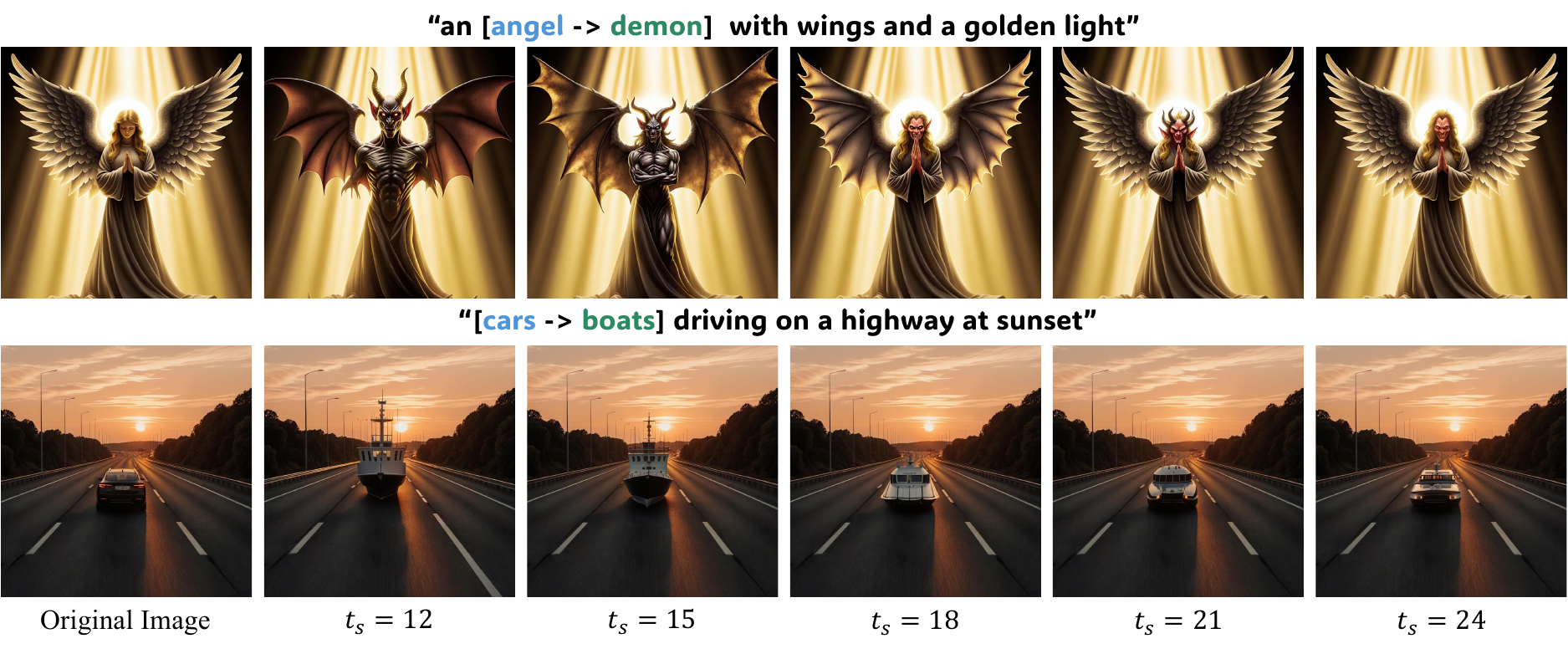}
\vspace{-0.6cm}
\caption{Effect of the $t_s$. Earlier branching permits stronger
changes but can disturb the background, whereas later branching
retains more source appearance and may leave the edit incomplete.}
\vspace{-0.1cm}
\label{fig:switch}
\end{figure}

\textbf{\textit{Effects of hyper-parameters.}}
We qualitatively examine the three core parameters: the switch step
$t_s$, the spatial threshold $\tau_{\mathrm{spa}}$, and the bitwise
threshold $\tau_{\mathrm{pow}}$.
Figure~\ref{fig:switch} shows the trade-off in choosing when to begin
editing. Branching too early introduces target semantics before a clear
source layout has emerged, allowing changes to spread into the
background. Branching too late better preserves the source structure
but leaves the intended transformation incomplete. This is consistent
with a more established visual state being harder to change and fewer
remaining steps for editable target refinement; the qualitative comparison alone
does not distinguish these effects. In Figure~\ref{fig:switch},
$t_s=18$ provides a suitable balance between editability and preservation.

\begin{figure}[htbp]
\centering
\includegraphics[width=\linewidth]{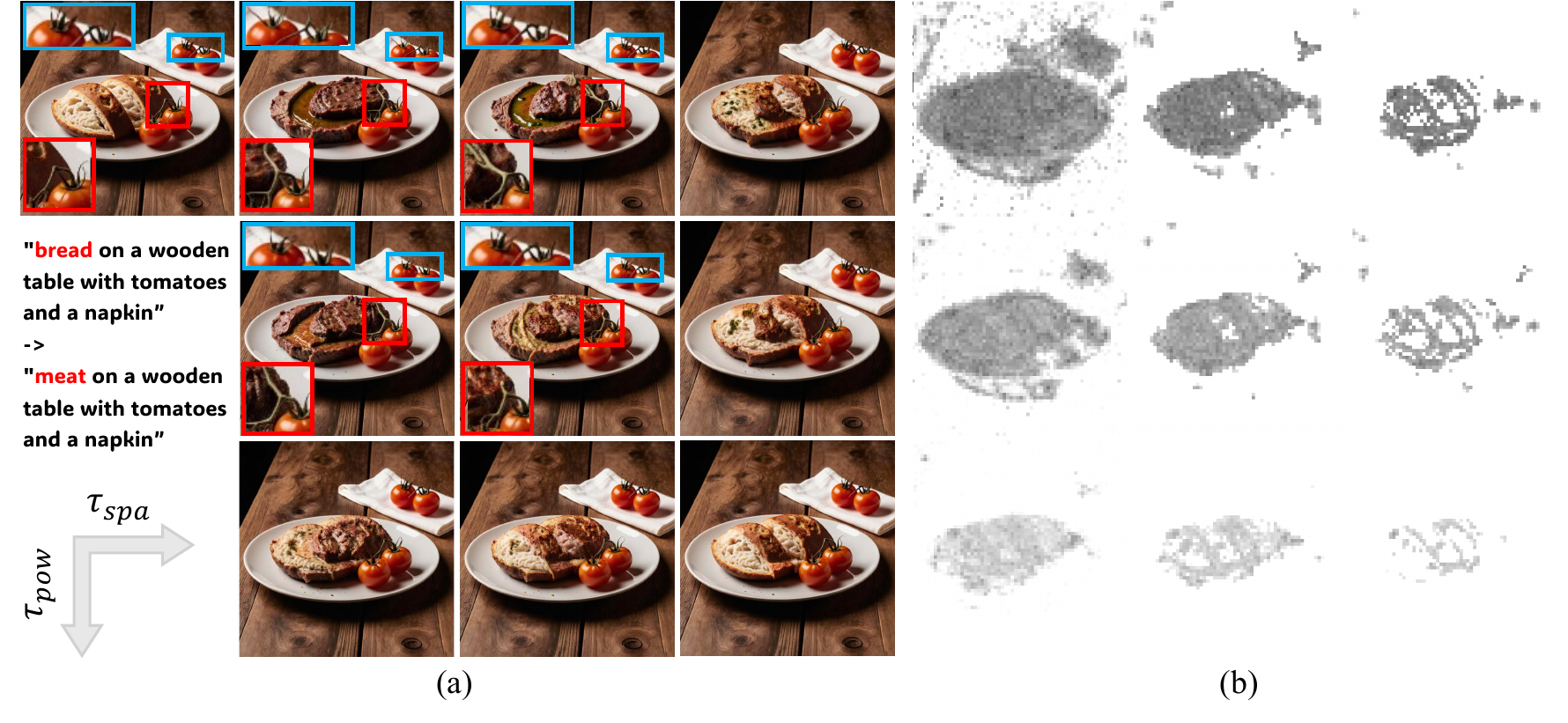}
\vspace{-0.8cm}
\caption{Joint effects of spatial and bitwise thresholds.
(a) Edited images. (b) Corresponding bit-density maps, where darker
pixels indicate more active bits. Columns use
$\tau_{\mathrm{spa}}=0.1, 0.2, 0.3$ from left to right;
rows use $\tau_{\mathrm{pow}}=0.12, 0.16, 0.20$
from top to bottom.}
\label{fig:parameter}
\vspace{-0.35cm}
\end{figure}

With the switch step fixed, Figure~\ref{fig:parameter} shows
how the two thresholds control spatial coverage and bitwise
edit strength.
Increasing $\tau_{\mathrm{spa}}$ from left to right
reduces the selected region in Figure~\ref{fig:parameter}~(b), leaving more bread
unchanged in Figure~\ref{fig:parameter}~(a).
Increasing $\tau_{\mathrm{pow}}$ from top to bottom reduces
the fraction of active bits, producing lighter density maps
and weaker edits.
The thresholds interact through refinement: bit selection
changes the editing state and the probability evidence used
for later spatial masks.
In these examples, stronger bitwise editing also expands
the mask, highlighting the balance between editing coverage
and background preservation.
Controlled quantitative comparisons are provided in
Appendix~\ref{app:parameter_sensitivity}.

\begin{wrapfigure}{r}{0.62\linewidth}
    \centering
    \vspace{-0.5cm}
    \includegraphics[width=\linewidth]{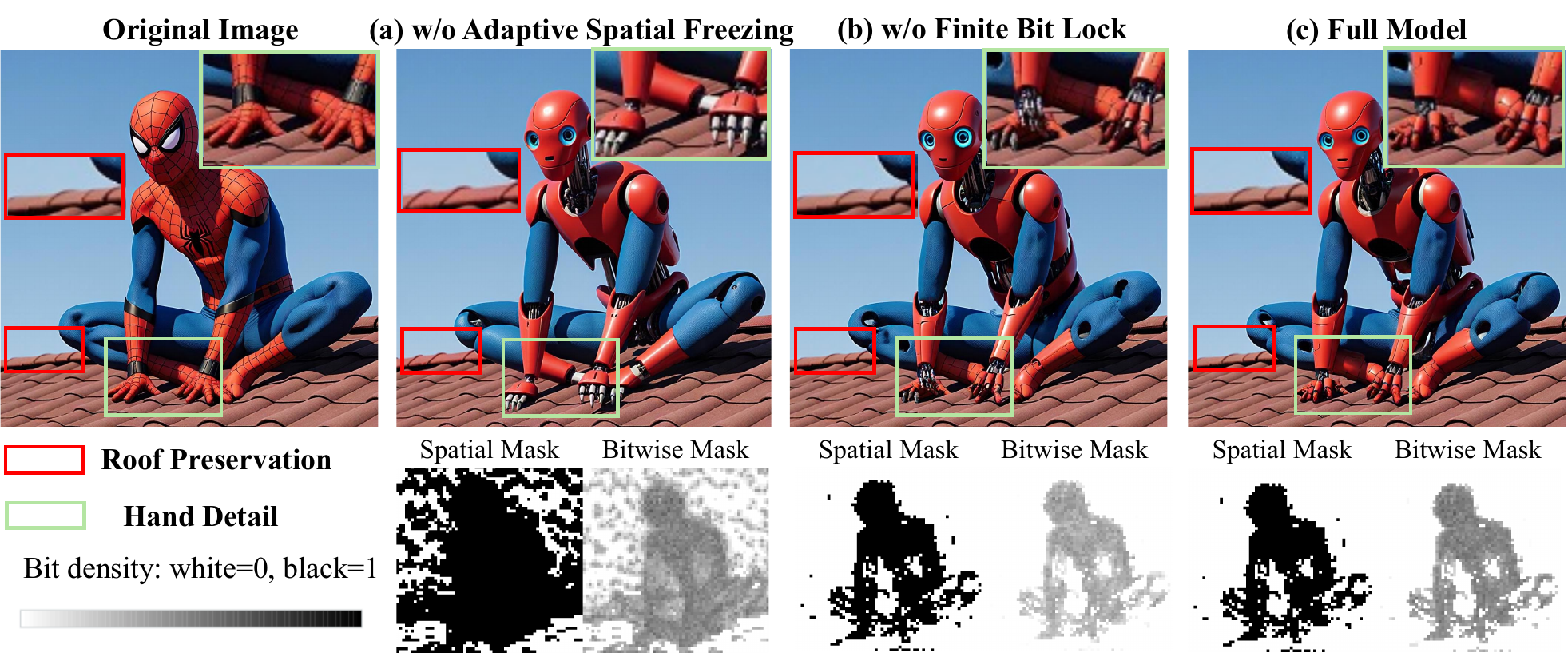}
    \vspace{-0.4cm}
    \caption{Qualitative mechanism ablation. Removing freezing edits the roof around the robot; removing bit locking leaves the fingers incompletely edited. Spatial masks and bit-density maps show the corresponding changes in editing support.}
    \label{fig:mechanism_ablation}
    \vspace{-0.1cm}
\end{wrapfigure}
\textbf{\textit{Effects of mask stabilization.}}
Figure~\ref{fig:mechanism_ablation} compares variants without
\textit{AdaSF} or \textit{FBL}, with other settings unchanged.
Without \textit{AdaSF}, the editing mask spreads into the
roof and background near the robot's hands, introducing
unwanted changes.
Without \textit{FBL}, the mask remains more localized,
but the fingers are not fully edited.
Together, the two mechanisms balance spatial control
with continued refinement of edited content.
Quantitative results are provided in
Appendix~\ref{app:mechanism_ablation}.

\vspace{-0.1cm}

\section{Conclusion}
\label{conclu}
In this paper, we presented \textbf{\textit{RefineEdit}},
a training-free framework that adapts GRN to prompt-to-prompt
image editing.
Starting from an intermediate source state, it uses aligned
bit probability differences to guide editing while preserving
unselected source coordinates.
It incorporates two stabilization techniques,
\textit{adaptive spatial freezing} and \textit{finite bit locking}.
\textit{Adaptive spatial freezing} limits unnecessary expansion of the editing region, while \textit{finite bit locking} keeps recently selected bits editable across refinement steps, reducing interruptions to the refinement of edited content. Experiments across nine PIE-Bench editing categories show strong preservation of unedited content and alignment with editing prompts.
These results support our central insight that
\textit{refinement is inherently editable}:
edit localization and content generation can evolve within
the same process. Exploring this insight beyond GRN could open new directions for training-free image editing.

\vspace{-0.1cm}

\clearpage

\bibliography{iclr2027_conference}
\bibliographystyle{iclr2027_conference}

\clearpage
\appendix
\renewcommand{\theHtable}{app.\arabic{table}}
\renewcommand{\theHfigure}{app.\arabic{figure}}
\tocheading{Appendix Contents}
\begingroup
\tocmain{A}{Implementation and Hyperparameter Settings}{app:refineedit_settings}
    \tocsub{A.1}{Fixed GRN Settings}{app:grn_settings}
    \tocsub{A.2}{RefineEdit Hyperparameters}{app:refineedit_settings}
    \tocsub{A.3}{Controlled Hyperparameter Comparisons}{app:parameter_sensitivity}
    \tocsub{A.4}{Mechanism Ablations}{app:mechanism_ablation}

\tocmain{B}{User Study}{app:user_study} 

\tocmain{C}{Evaluation Details}{app:evaluation}
    \tocsub{C.1}{Dataset}{app:dataset}
    \tocsub{C.2}{Evaluation Metrics}{app:metrics}
    \tocsub{C.3}{Compared Methods}{app:baselines}
    \tocsub{C.4}{Cross-Dataset Evaluation on EditEval v2}{app:editeval}

\tocmain{D}{Algorithm of RefineEdit}{app:algorithm}

\tocmain{E}{Additional Qualitative Results}{app:qualitative_results}
\endgroup

\section{Implementation and Hyperparameter Settings}
\label{app:refineedit_settings}

\subsection{Fixed GRN Settings}
\label{app:grn_settings}
We use the pretrained GRN-2B~\citep{grn} text-to-image model with the frozen HBQ
tokenizer and UMT5-XXL text encoder. All network weights remain fixed.
Table~\ref{tab:grn_settings} lists the shared inference settings.
Images are generated at $1024\times1024$ resolution. The corresponding
HBQ code has $64\times64$ spatial positions and 256 bits per position,
obtained from 64 latent channels and four quantization rounds.
Generation starts from random binary labels, with each bit sampled
as 0 or 1 with equal probability. The two branches share this initial
random code, which is reused whenever a coordinate returns to a random
label. We use separate random generators for source and editing
sampling, seeded with 42 and 43, respectively.

\begin{table}[htbp]
\centering
\small
\caption{Fixed GRN settings used for \textbf{\textit{RefineEdit}}. These settings are
shared across editing categories.}
\label{tab:grn_settings}
\renewcommand{\arraystretch}{1.12}
\begin{tabular*}{\linewidth}{@{\extracolsep{\fill}}ll@{}}
\toprule
\textbf{Setting} & \textbf{Value} \\
\midrule
Generative backbone & GRN-2B, T2I checkpoint 251600 \\
Visual tokenizer & HBQ, 64 channels, 4 quantization rounds \\
Tokenizer checkpoint & Image/video tokenizer dated 2026-06-26 \\
Text encoder & UMT5-XXL \\
Generation resolution & $1024\times1024$ \\
Spatial code grid & $64\times64$ \\
Refinement iterations & 50, indexed from 0 to 49 \\
Classifier-free guidance scale & 3.0 \\
Sampling temperature & 1.1 \\
Bit sampling & Multinomial sampling over $\{0,1\}$ \\
Prediction-retention schedule & Cosine schedule, maximum ratio 0.95 \\
Schedule shift & 1.0 \\
Source / editing random seeds & 42 / 43 \\
Negative prompt & Empty string \\
Inference autocast precision & bfloat16 \\
Additional attention control & None \\
\bottomrule
\end{tabular*}
\end{table}

Both branches follow the same global refinement schedule and finish
at step 49. For the input to zero-based step $t$, the nominal
prediction-retention ratio is
\begin{equation}
\lambda_t=0.95\left[1-\cos\left(\frac{\pi t}{2\cdot49}\right)\right],
\qquad t=0,\ldots,49.
\label{eq:appendix_grn_schedule}
\end{equation}
The selection mask is sampled at each step, so its realized prediction
fraction can differ slightly from this nominal ratio. The implementation
uses the realized source fraction for the shared progress embedding.
At the final step, we decode the sampled predictions after applying
source/edit routing, rather than decoding the intermediate state that
still contains random labels. We do not add refinement steps or restart
the noise schedule after editing begins. A switch step $t_s$ therefore
leaves $50-t_s$ editing iterations, including the branching step itself.

\subsection{\textbf{\textit{RefineEdit}} Hyperparameters}
\label{app:refineEdit_setting}
We select the switch step $t_s$, spatial threshold
$\tau_{\mathrm{spa}}$, and bitwise threshold
$\tau_{\mathrm{pow}}$ for each editing category.
Table~\ref{tab:category_settings} lists the configurations
used in the main quantitative comparison.
All images within a category share the same configuration.

\begin{table}[t]
\centering
\small
\caption{Category-specific settings used in the quantitative
comparison. All categories use $K=4$ and
$\tau_{\mathrm{frz}}=2\tau_{\mathrm{spa}}$.}
\label{tab:category_settings}
\renewcommand{\arraystretch}{1.15}
\begin{tabular*}{\linewidth}{@{\extracolsep{\fill}}llrrrr@{}}
\toprule
\textbf{Subset} & \textbf{Editing category} & \textbf{Images}
& $t_s$ & $\tau_{\mathrm{spa}}$ & $\tau_{\mathrm{pow}}$ \\
\midrule
1 & Object replacement & 80 & 15 & 0.013 & 0.20 \\
2 & Object addition & 80 & 15 & 0.016 & 0.10 \\
3 & Object removal & 80 & 10 & 0.020 & 0.16 \\
4 & Content modification & 40 & 12 & 0.016 & 0.18 \\
5 & Pose modification & 40 & 12 & 0.018 & 0.16 \\
6 & Color modification & 40 & 11 & 0.020 & 0.16 \\
7 & Material modification & 40 & 16 & 0.020 & 0.20 \\
8 & Background modification & 80 & 10 & 0.012 & 0.16 \\
9 & Style transfer & 80 & 6 & 0.016 & 0.14 \\
\bottomrule
\end{tabular*}
\end{table}

\subparagraph{Fixed mask stabilization.}
We set $K=4$ for all categories. A bit selected in any of the latest
four refinement steps remains eligible for editing, but its binary
value can continue to change. For adaptive spatial freezing, we fix
the response multiplier to 2, giving
$\tau_{\mathrm{frz}}=2\tau_{\mathrm{spa}}$.
This fixes the multiplier, not the absolute threshold across categories.
If the mean excess spatial response at the switch step reaches this
threshold, we freeze the spatial mask immediately. Otherwise, the mask
remains dynamic through the final step. These mechanisms are evaluated
by removing each one separately, rather than by sweeping their settings.

\subsection{Controlled Hyperparameter Comparisons}
\label{app:parameter_sensitivity}
Table~\ref{tab:parameter_sensitivity} reports controlled comparisons for the three core parameters. The switch-step comparison uses the object-replacement subset, while the spatial and bitwise threshold comparisons use the color-modification subset. Each block varies only the indicated parameter. The other two parameters, random seeds, GRN settings, $K=4$, and the adaptive-freezing
rule remain fixed. All entries are taken from completed grid runs,
with the means verified against their per-image evaluation records.
These are slices of the existing search, not additional runs.

\subparagraph{Switch step.}
Later branching improves background preservation in this comparison,
while both CLIP scores decrease. This supports a trade-off between
retaining source structure and allowing target changes.
Because the endpoint stays fixed, changing $t_s$ changes both the
inherited source state and the number of editing iterations. These
results do not isolate either factor as the sole cause.

\subparagraph{Spatial threshold.}
On the color-modification subset, increasing $\tau_{\mathrm{spa}}$
improves background preservation but lowers edited-region CLIP.
This illustrates the trade-off between restricting spatial
selection and allowing the requested semantic change.
The comparison uses the full AdaSF rule, so
$\tau_{\mathrm{frz}}$ scales with $\tau_{\mathrm{spa}}$.

\subparagraph{Bitwise threshold.}
Increasing $\tau_{\mathrm{pow}}$ restricts the bits eligible
for editing at each selected position.
In this comparison, all reported background metrics improve,
but both whole-image and edited-region CLIP decrease.
Stricter bit selection therefore favors source preservation
at the expense of semantic editing.

\subsection{Mechanism Ablations}
\label{app:mechanism_ablation}

We evaluate \textit{AdaSF} and \textit{FBL} on the object-replacement subset by
disabling one mechanism at a time.
Without \textit{AdaSF}, the spatial mask is updated at every step,
with \textit{FBL} still enabled.
Without \textit{FBL}, only the current step's candidate bits remain active,
equivalent to $K=1$, with \textit{AdaSF} still enabled.
All other settings remain unchanged.
Figure~\ref{fig:mechanism_ablation} and
Table~\ref{tab:mechanism_ablation} present the qualitative
and quantitative comparisons, respectively.

Removing \textit{FBL} improves background preservation but lowers both
CLIP scores, indicating that retaining recent bit activations
supports semantic editing.
Removing \textit{AdaSF} improves CLIP scores at the cost of background
preservation and structural consistency.
These results illustrate their complementary roles:
\textit{AdaSF} restricts unwanted spatial expansion, and \textit{FBL} sustains
editing activity across refinement steps.

\begin{table}[h]
\centering
\small
\caption{Representative controlled parameter comparisons.
The switch-step comparison uses the object-replacement subset,
and the threshold comparisons use the color-modification subset.
Only one input hyperparameter varies within each block,
with AdaSF and FBL enabled.
Background metrics are evaluated outside the reference edit mask.
Higher PSNR, SSIM, and CLIP and lower LPIPS are better.}
\label{tab:parameter_sensitivity}
\setlength{\tabcolsep}{3pt}
\renewcommand{\arraystretch}{1.15}
\begin{tabular*}{\linewidth}{@{\extracolsep{\fill}}rrrrrrrr@{}}
\toprule
\multicolumn{3}{c}{\textbf{Hyperparameters}}
& \multicolumn{3}{c}{\textbf{Background Preservation}}
& \multicolumn{2}{c}{\textbf{CLIP Similarity}} \\
\cmidrule(lr){1-3}\cmidrule(lr){4-6}\cmidrule(lr){7-8}
$t_s$ & $\tau_{\mathrm{spa}}$ & $\tau_{\mathrm{pow}}$
& PSNR$\uparrow$ & LPIPS$\downarrow$ & SSIM$\uparrow$
& Whole$\uparrow$ & Edited$\uparrow$ \\
\midrule
\multicolumn{8}{l}{\textit{Varying the switch step}} \\
15 & 0.016 & 0.14 & 26.08 & 0.042 & 0.939 & 26.04 & 20.12 \\
18 & 0.016 & 0.14 & 29.18 & 0.027 & 0.959 & 25.05 & 19.33 \\
20 & 0.016 & 0.14 & 31.14 & 0.019 & 0.969 & 24.34 & 18.57 \\
\midrule
\multicolumn{8}{l}{\textit{Varying the spatial threshold}} \\
14 & 0.010 & 0.12 & 29.11 & 0.031 & 0.948 & 26.73 & 21.95 \\
14 & 0.014 & 0.12 & 33.25 & 0.023 & 0.965 & 26.54 & 21.55 \\
14 & 0.020 & 0.12 & 36.49 & 0.018 & 0.976 & 26.42 & 21.35 \\
\midrule
\multicolumn{8}{l}{\textit{Varying the bitwise threshold}} \\
13 & 0.010 & 0.09 & 26.92 & 0.037 & 0.937 & 27.36 & 22.46 \\
13 & 0.010 & 0.12 & 28.67 & 0.031 & 0.950 & 27.07 & 22.24 \\
13 & 0.010 & 0.16 & 31.11 & 0.023 & 0.965 & 26.87 & 21.69 \\
\bottomrule
\end{tabular*}
\end{table}

\vspace{-0.5cm}

\begin{table}[h]
\centering
\small
\caption{Mechanism ablations on the object-replacement subset.
Each variant disables one mechanism while keeping the other
settings unchanged. Arrows indicate the better direction.}
\label{tab:mechanism_ablation}
\setlength{\tabcolsep}{3pt}
\renewcommand{\arraystretch}{1.2}
\begin{tabular*}{\linewidth}{@{\extracolsep{\fill}}lrrrrrr@{}}
\toprule
\multirow{2}{*}{\textbf{Method}}
& \multicolumn{1}{c}{\textbf{Structure}}
& \multicolumn{3}{c}{\textbf{Background Preservation}}
& \multicolumn{2}{c}{\textbf{CLIP Similarity}} \\
\cmidrule(lr){2-2}\cmidrule(lr){3-5}\cmidrule(lr){6-7}
& \multicolumn{1}{c}{\textbf{Distance}$\downarrow$}
& \multicolumn{1}{c}{\textbf{PSNR}$\uparrow$}
& \multicolumn{1}{c}{\textbf{LPIPS}$\downarrow$}
& \multicolumn{1}{c}{\textbf{SSIM}$\uparrow$}
& \multicolumn{1}{c}{\textbf{Whole}$\uparrow$}
& \multicolumn{1}{c}{\textbf{Edited}$\uparrow$} \\
\midrule
Full model
& 0.028 & 30.04 & 0.044 & 0.932 & 26.48 & 23.17 \\
w/o Finite Bit Locking
& 0.028 & 31.04 & 0.041 & 0.938 & 26.34 & 23.01 \\
w/o Adaptive Spatial Freezing
& 0.033 & 28.76 & 0.050 & 0.920 & 26.59 & 23.44 \\
\bottomrule
\end{tabular*}
\end{table}

\section{User Study}
\label{app:user_study}
\begin{figure}[h]
    \centering
    \includegraphics[width=\linewidth]{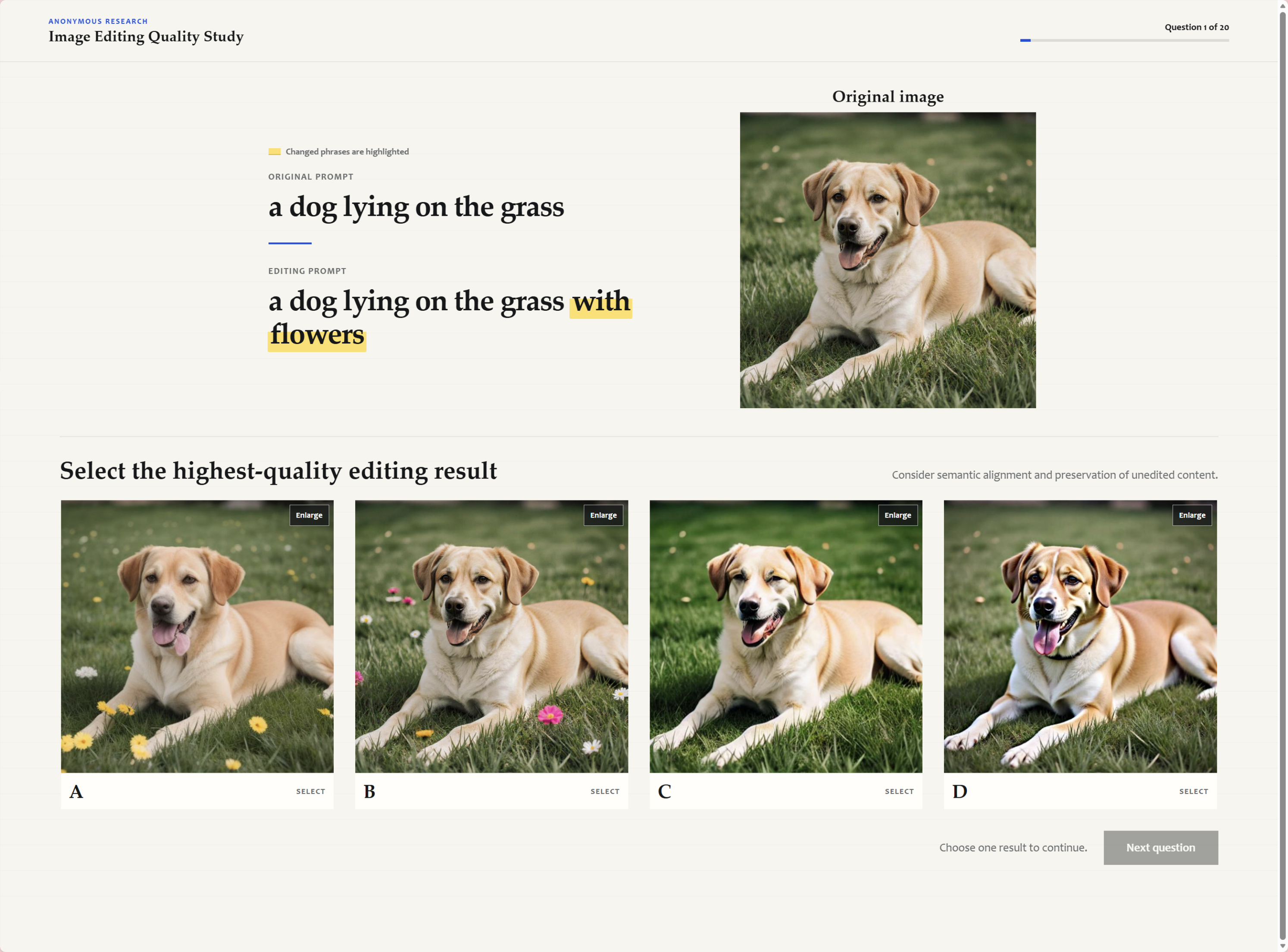}
    \caption{User study interface.}
    \label{fig:user-study-ui}
    \vspace{-0.35cm}
\end{figure}

\noindent
\begin{minipage}[t]{0.49\linewidth}
    \vspace{0pt}
    We conducted a user study with 20 participants to evaluate
    human preferences for our editing results against those of
    LEDits++, FlowEdit, and ChordEdit. The study comprised 20 sets
    of editing results (400 votes in total) randomly sampled from a pool spanning all
    editing categories. As illustrated in Fig.~\ref{fig:user-study-ui}, participants
    were shown four candidate results with the method names hidden
    and asked to select the best one. As shown in
    Fig.~\ref{fig:user-study-result}, our method received a higher
    proportion of preference votes than each baseline.
\end{minipage}\hfill
\begin{minipage}[t]{0.38\linewidth}
    \vspace{0pt}
    \centering
    \includegraphics[width=0.9\linewidth]{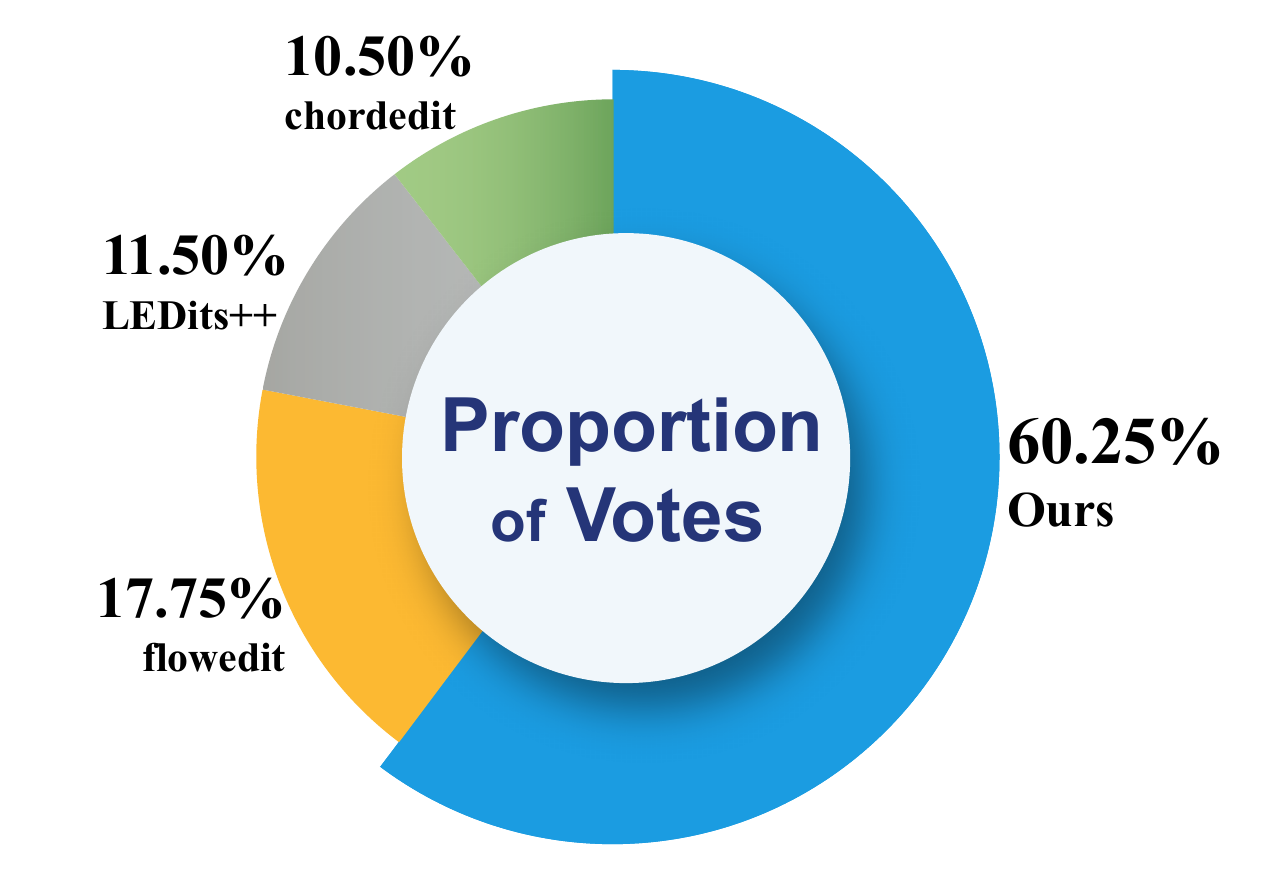}
    \captionof{figure}{Human preference results from the user study.}
    \label{fig:user-study-result}
\end{minipage}
\par

\section{Evaluation Details}
\label{app:evaluation}

\subsection{Dataset}
\label{app:dataset}

We construct a generation-to-editing evaluation set from PIE-Bench,
containing 560 prompt pairs across nine editing categories.
Each pair consists of a source prompt and an editing prompt that
specifies the desired change. Table~\ref{tab:dataset_categories}
summarizes the categories and their sample counts.

For each source prompt, we use GRN to generate a source image at
$1024 \times 1024$ resolution. \textbf{\textit{RefineEdit}} edits the corresponding
generation trajectory, while all baselines receive the same source
image. This shared source ensures that content preservation is
evaluated against the same reference across methods. The source
image serves as the reference for preservation, rather than as a
ground-truth edited target.

We use Grounded-SAM to obtain source-aligned region annotations
for evaluation. The reference editing masks specify which regions
may change and are distinct from the masks predicted by \textbf{\textit{RefineEdit}}.
They are used only for evaluation and are not provided to our method
during inference.

Before evaluation, source and edited images are resized to
$512 \times 512$ using interpolation. Reference masks are
decoded at their encoded resolution and resized using nearest-neighbor
interpolation to preserve their binary values.
This protocol standardizes the evaluation resolution across methods.

\begin{table}[t]
    \centering
    \small
    \setlength{\tabcolsep}{4pt}
    \caption{
    Composition of our PIE-Bench-derived evaluation set.
    Examples summarize the edits illustrated in
    Fig.~\ref{fig:piebench_examples}.
    }
    \label{tab:dataset_categories}
    \begin{tabular}{@{}p{0.22\linewidth}r p{0.66\linewidth}@{}}
        \toprule
        Category & Samples & Description and example \\
        \midrule
        Object replacement & 80 &
        Replace an object with another category,
        such as cars with boats. \\

        Object addition & 80 &
        Introduce a new object,
        such as flowers on the grass beside a dog. \\

        Object removal & 80 &
        Remove an existing object,
        such as a rose held by a teddy bear. \\

        Content modification & 40 &
        Modify an object's attributes or internal appearance,
        such as adding black and yellow stripes to a tie. \\

        Pose modification & 40 &
        Change an object's pose, action, or orientation,
        such as making a girl look toward the camera. \\

        Color modification & 40 &
        Change an object's color,
        such as turning a white kitten yellow. \\

        Material modification & 40 &
        Change an object's material appearance,
        such as giving a person a golden appearance. \\

        Background modification & 80 &
        Change the surrounding scene,
        such as replacing a beach view with a garden. \\

        Style transfer & 80 &
        Change the image's overall visual style,
        such as converting a photograph into a watercolor painting. \\
        \bottomrule
    \end{tabular}
\end{table}

\begin{figure}[p]
    \centering
    \includegraphics[
        width=\linewidth,
        height=\textheight,
        keepaspectratio
    ]{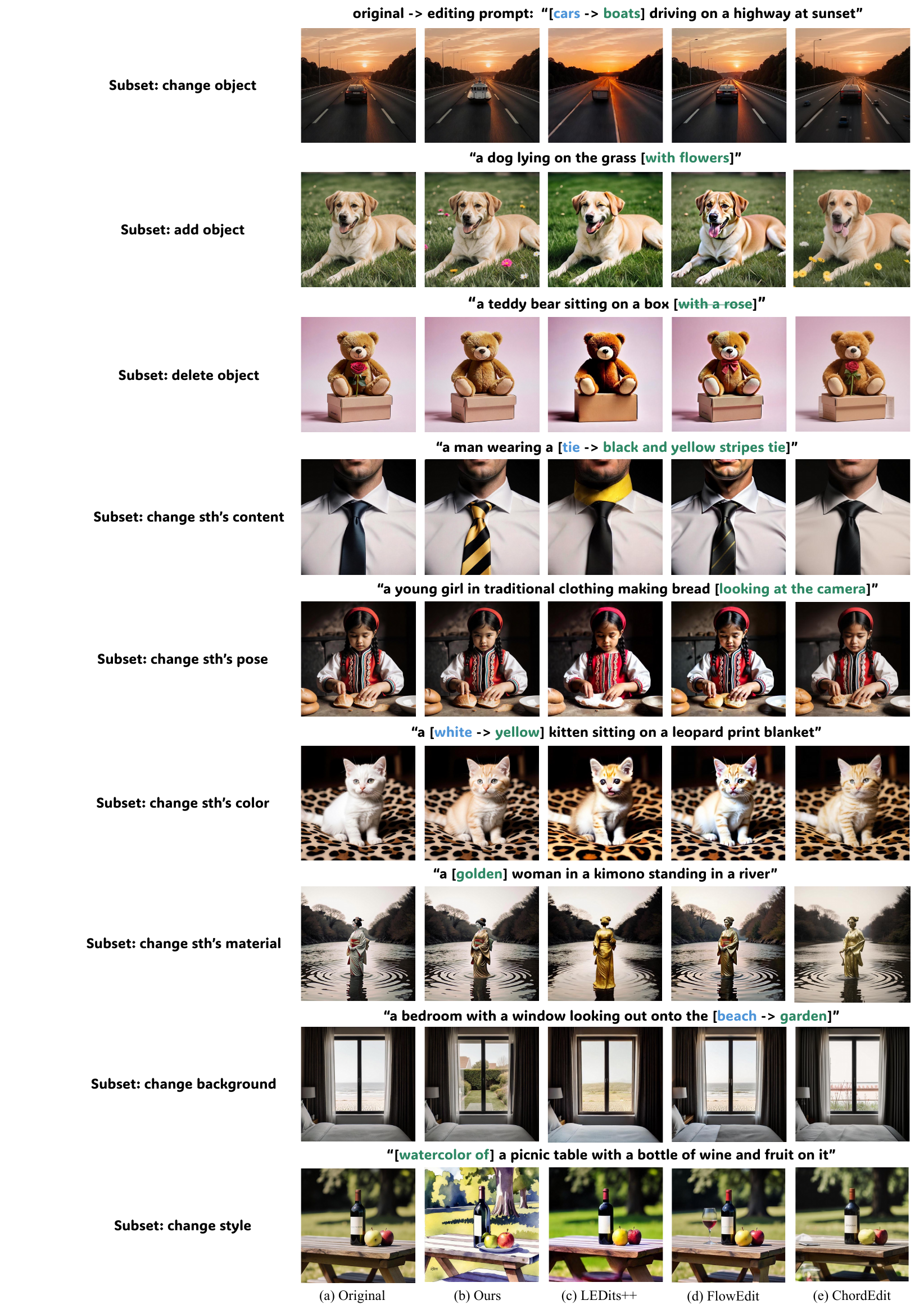}
    \caption{
    Examples from the nine editing categories.
    Each row shows a source image and the corresponding results
    from \textbf{\textit{RefineEdit}}, LEDits++, FlowEdit, and ChordEdit.
    Prompt annotations indicate the content to replace, add, or remove.
    }
    \label{fig:piebench_examples}
\end{figure}

Figure~\ref{fig:piebench_examples} presents one example from each
category, including the prompt change, source image, and outputs
from \textbf{\textit{RefineEdit}} and representative baselines. These examples cover
both localized object edits and broader changes to the scene or style.

\subsection{Evaluation Metrics}
\label{app:metrics}

We evaluate structural consistency, preservation of unedited content,
and alignment with the editing prompt.
Let $I^s$ and $I^e$ denote the source and edited RGB images,
and let $c^e$ denote the editing prompt.
Let $M \in \{0,1\}^{H \times W}$ be the reference editing mask,
where $M=1$ indicates an editable region.
The images used to evaluate preservation are
\begin{equation}
    U=(1-M)\odot I^s,
    \qquad
    V=(1-M)\odot I^e,
    \label{eq:eval_preservation_inputs}
\end{equation}
where the mask is broadcast across RGB channels.
Thus, editable regions are set to zero in both images.

Although we refer to these measures as background preservation,
they evaluate all content outside the editing mask.
For background modification, this preserved content may instead
be the foreground subject.

\subparagraph{Structure Distance.~\citep{tumanyan2022splicing}}
Structure Distance measures changes in the relationships between
image tokens using a pretrained DINO ViT-B/8.
Let $k_i(I)$ denote the concatenated attention-key features of
token $i$ from the final Transformer layer, following the
preprocessing used by our evaluation implementation.
The token self-similarity matrix is
\begin{equation}
    A_{ij}(I)
    =
    \frac{
        k_i(I)^\top k_j(I)
    }{
        \max\!\left(
            \|k_i(I)\|_2\|k_j(I)\|_2,\epsilon
        \right)
    },
    \label{eq:eval_dino_similarity}
\end{equation}
where $\epsilon$ ensures numerical stability.
Structure Distance is the mean squared difference between the
source and edited self-similarity matrices:
\begin{equation}
    D_{\mathrm{structure}}
    =
    \frac{1}{T^2}
    \sum_{i=1}^{T}\sum_{j=1}^{T}
    \left(
        A_{ij}(I^s)-A_{ij}(I^e)
    \right)^2,
    \label{eq:eval_structure_distance}
\end{equation}
where $T$ is the number of image tokens, including the class token
used by the implementation.
This metric is computed on the full images.
Lower values indicate stronger structural consistency.

\subparagraph{Mean Squared Error (MSE).~\citep{wang2009mean}}
MSE measures pixel differences in the masked preservation images.
With RGB values normalized to $[0,1]$, it is computed as
\begin{equation}
    \operatorname{MSE}
    =
    \frac{1}{3HW}
    \sum_{h=1}^{H}
    \sum_{w=1}^{W}
    \sum_{c=1}^{3}
    \left(U_{hwc}-V_{hwc}\right)^2.
    \label{eq:eval_mse}
\end{equation}
Following the evaluation implementation, the denominator is the
size of the full RGB image, rather than the number of unmasked
coordinates. Lower values indicate better preservation.

\subparagraph{Peak Signal-to-Noise Ratio (PSNR).~\citep{wang2004image}}
PSNR expresses pixel reconstruction fidelity on a logarithmic scale.
For images normalized to $[0,1]$, it is
\begin{equation}
    \operatorname{PSNR}
    =
    10\log_{10}\left(\frac{1}{\operatorname{MSE}}\right),
    \label{eq:eval_psnr}
\end{equation}
where MSE is defined in Eq.~\ref{eq:eval_mse}.
Higher PSNR indicates better preservation of the source content.

\subparagraph{Learned Perceptual Image Patch Similarity (LPIPS).~\citep{zhang2018unreasonable}}
LPIPS measures perceptual differences using pretrained visual
features. We use the SqueezeNet implementation.
Let $\phi_\ell(U)_{hw}$ and $\phi_\ell(V)_{hw}$ be the
channel-normalized features at layer $\ell$ and spatial position
$(h,w)$. LPIPS is
\begin{equation}
    \operatorname{LPIPS}
    =
    \sum_{\ell}
    \frac{1}{H_\ell W_\ell}
    \sum_{h,w}
    w_\ell^\top
    \left(
        \phi_\ell(U)_{hw}
        -
        \phi_\ell(V)_{hw}
    \right)^{\odot 2},
    \label{eq:eval_lpips}
\end{equation}
where $w_\ell$ contains learned channel weights and
$\odot 2$ denotes element-wise squaring.
The masked images are mapped from $[0,1]$ to $[-1,1]$
before LPIPS preprocessing.
The metric is evaluated on the complete masked images,
without additional normalization by the preserved area.
Lower values indicate greater perceptual similarity.

\subparagraph{Structural Similarity Index Measure (SSIM).~\citep{wang2004image}}
SSIM compares local luminance, contrast, and structure.
For corresponding local windows $u$ and $v$ from $U$ and $V$,
\begin{equation}
    \operatorname{SSIM}(u,v)
    =
    \frac{
        (2\mu_u\mu_v+C_1)
        (2\sigma_{uv}+C_2)
    }{
        (\mu_u^2+\mu_v^2+C_1)
        (\sigma_u^2+\sigma_v^2+C_2)
    },
    \label{eq:eval_ssim}
\end{equation}
where $\mu_u$ and $\mu_v$ are local means,
$\sigma_u^2$ and $\sigma_v^2$ are local variances,
and $\sigma_{uv}$ is their covariance.
For the unit intensity range, we use
$C_1=0.01^2$ and $C_2=0.03^2$.
Local statistics use an $11 \times 11$ Gaussian window with
standard deviation $1.5$.
The final score averages the local SSIM values over the masked
images and RGB channels. Higher values indicate better preservation.

\subparagraph{CLIP Similarity.~\citep{radford2021learning}}
We measure semantic alignment using CLIP ViT-L/14.
Let $f_I$ and $f_T$ denote its image and text encoders,
including their respective preprocessing.
The image-text score is
\begin{equation}
    C(I,c)
    =
    100\max\left(
        0,\,
        \frac{
            f_I(I)^\top f_T(c)
        }{
            \|f_I(I)\|_2\|f_T(c)\|_2
        }
    \right).
    \label{eq:eval_clip}
\end{equation}
We report alignment with the editing prompt for both the full image
and the edited region:
\begin{equation}
    \operatorname{CLIP}_{\mathrm{tgt}}
    =
    C(I^e,c^e),
    \qquad
    \operatorname{CLIP}_{\mathrm{edit}}
    =
    C(M\odot I^e,c^e).
    \label{eq:eval_clip_variants}
\end{equation}
For edited-region CLIP, pixels outside the reference editing mask
are set to black before CLIP preprocessing; the image is not
cropped to the mask's bounding box.
Both scores use the complete editing prompt.
Higher values indicate stronger target alignment.

The evaluator additionally records source-image alignment,
$C(I^s,c^s)$, where $c^s$ is the source prompt.
This score describes the generated source image and is not an
editing-performance measure.

\subparagraph{Aggregation.}
Metrics are computed per image and averaged over finite
values using the same rule for all methods.
For metric $m$, let $\mathcal{V}_m$ contain samples with
finite scores. The reported mean is
\begin{equation}
    \overline{m}
    =
    \frac{1}{|\mathcal{V}_m|}
    \sum_{i\in\mathcal{V}_m}m_i.
    \label{eq:eval_aggregation}
\end{equation}
Each valid image receives equal weight, rather than each
category.
Undefined scores and infinite PSNR values are excluded,
not replaced by zero.
Valid counts can therefore differ across metrics and methods.
PSNR is averaged per image, rather than computed from the
pooled MSE.

\subsection{Compared Methods}
\label{app:baselines}

We compare \textbf{\textit{RefineEdit}} with nine training-free editing methods
based on pretrained diffusion or rectified-flow models.
We group the baselines by the generative backbones used in our
evaluation. Diffusion methods use SD 1.4, SD 1.5, or SD-Turbo,
while flow methods use FLUX.1-dev.

\paragraph{Diffusion Editing Methods}

\subparagraph{Prompt-to-Prompt (P2P).~\citep{p2p}}
Prompt-to-Prompt controls the cross-attention maps that connect
text tokens to image regions.
By replacing, retaining, or reweighting these maps during generation,
it introduces changes specified by the editing prompt while
preserving the source layout.

\subparagraph{MasaCtrl.~\citep{masactrl}}
MasaCtrl replaces diffusion self-attention with mutual self-attention,
allowing the editing branch to retrieve related content and textures
from the source branch.
Masks derived from cross-attention help distinguish foreground
from background and reduce incorrect feature matching.
This supports changes in pose and shape while preserving
the source object's appearance.

\subparagraph{Pix2Pix-Zero.~\citep{parmar2023zero}}
Pix2Pix-Zero represents a semantic edit as a direction in the
text-embedding space.
It uses cross-attention guidance to retain the source attention
maps during editing, helping preserve content outside the
intended change.
The method requires no task-specific training.

\subparagraph{Plug-and-Play (PnP).~\citep{pnp}}
PnP extracts spatial features and self-attention information
from the source diffusion trajectory and injects them into
generation conditioned on the editing prompt.
These features preserve the source layout while the editing
prompt guides changes in appearance.
The method requires no additional training or fine-tuning.

\subparagraph{PnP with Direct Inversion (PnP-DirInv).~\citep{directinversionandpiebench}}
This baseline combines PnP with Direct Inversion, which separates
source reconstruction from editing.
The source branch preserves the input, while the editing branch
introduces the requested changes.
PnP provides additional structure control through feature injection.
We evaluate this combination separately from standard PnP.

\subparagraph{LEDits++.~\citep{leditspp}}
LEDits++ combines an efficient, optimization-free inversion
procedure with semantic editing guidance.
Its implicit masks restrict changes to relevant image regions.
The framework also supports multiple simultaneous edits.

\subparagraph{ChordEdit.~\citep{chordedit}}
ChordEdit formulates editing as transport between distributions
defined by the source and editing prompts.
Using dynamic optimal transport, it derives a smoothed,
low-energy editing field that reduces trajectory variance.
This enables editing through a single large integration step
without additional training.
We evaluate ChordEdit with the SD-Turbo backbone.

\paragraph{Flow Editing Methods}

\subparagraph{FlowEdit.~\citep{flowedit}}
FlowEdit constructs an ordinary differential equation that
directly transports images between distributions defined by
the source and editing prompts.
It combines the corresponding model fields to form an editing
trajectory without first recovering a noise latent.
The method requires neither inversion nor test-time optimization.

\subparagraph{RF-Inversion.~\citep{rfinversion}}
RF-Inversion supports reconstruction and editing with pretrained
rectified-flow models.
It derives a controlled inversion field using a linear quadratic
regulator and relates the resulting dynamics to a rectified
stochastic differential equation.
The controlled trajectory enables editing while retaining
source content, without additional model training.

\subsection{Cross-Dataset Evaluation on EditEval v2}
\label{app:editeval}

\paragraph{\textit{Evaluation setup.}}
We conduct an additional evaluation on EditEval v2~\citep{imageEditingSurvey} to assess
whether the category-specific settings selected on PIE-Bench
transfer to another dataset without retuning.
EditEval v2 contains 150 source and editing prompt pairs
across seven categories: object replacement, object addition,
object removal, action change, texture change, background
change, and style change.
Each category uses the corresponding PIE-Bench configuration
in Table~\ref{tab:category_settings}.
Action and texture changes use the settings for pose and
material modification, respectively.
All other GRN settings remain unchanged.

We follow the generation-to-editing protocol described in
Appendix~\ref{app:dataset}.
GRN generates a source image from each source prompt, and
all methods edit this same image using the editing prompt.
Thus, the evaluation uses EditEval prompt pairs with source
images produced by GRN, rather than the original benchmark
photographs.
Image preprocessing, regional evaluation, and metric
aggregation follow the same protocol as our PIE-Bench
experiments.

\paragraph{\textit{Results.}}
Table~\ref{tab:editeval_comparison} highlights the balance
between content preservation and semantic editing.
\textbf{\textit{RefineEdit}} achieves the best PSNR, LPIPS,
and SSIM, together with the second-best whole-image and
edited-region CLIP scores among the evaluated methods.
Although LEDits++ achieves higher CLIP scores, this advantage
is accompanied by substantially poorer performance on all
four background-preservation metrics.
FlowEdit obtains lower Structure Distance and MSE, but
shows weaker alignment with editing prompts and lower
preservation performance on PSNR, LPIPS, and SSIM.
In comparison, \textbf{\textit{RefineEdit}} combines
competitive prompt alignment with leading performance
on three complementary preservation metrics.
This balance is maintained on EditEval using the
PIE-Bench settings without retuning, supporting their
transferability across datasets.

\begin{table*}[htbp]
\centering
\caption{
Quantitative comparison across seven editing categories of
EditEval v2 under our generation-to-editing protocol.
\textbf{\textit{RefineEdit}} reuses the corresponding PIE-Bench
settings without retuning.
Methods are grouped by generation paradigm.
\textbf{Bold} and \underline{underlined} values indicate
the best and second-best results, respectively.
Cell colors indicate normalized performance within each metric,
from \textcolor{heatblue}{blue} for worst through white to
\textcolor{heatpink}{pink} for best.
}
\label{tab:editeval_comparison}

\setlength{\tabcolsep}{6pt}
\fontsize{7.4}{8.4}\selectfont

\resizebox{\linewidth}{!}{%
\begin{tabular}{ll c cccc cc}
\toprule
\multirow{2}{*}{\textbf{Method}}
& \multirow{2}{*}{\textbf{Model}}
& \multicolumn{1}{c}{\textbf{Struct.}}
& \multicolumn{4}{c}{\textbf{Background Preservation}}
& \multicolumn{2}{c}{\textbf{CLIP Sim.}} \\[-0.15em]
& &
\textbf{Dist.}$\downarrow$
& \textbf{PSNR}$\uparrow$
& \textbf{LPIPS}$\downarrow$
& \textbf{MSE}$\downarrow$
& \textbf{SSIM}$\uparrow$
& \textbf{Whole}$\uparrow$
& \textbf{Edited}$\uparrow$ \\
\cmidrule(lr){3-3}
\cmidrule(lr){4-7}
\cmidrule(lr){8-9}
\midrule

\rowcolor{datasetgray}
\multicolumn{9}{@{}l}{%
\hspace{0.25em}\textbf{\textit{Diffusion Methods}}} \\[-0.1em]

P2P & SD 1.4
& \hval{0.049}{0.019}{0.079}{0}
& \hval{17.97}{16.90}{24.60}{1}
& \hval{0.168}{0.065}{0.228}{0}
& \hval{0.0198}{0.0083}{0.0278}{0}
& \hval{0.701}{0.610}{0.892}{1}
& \hval{23.89}{22.85}{25.74}{1}
& \hval{20.37}{19.58}{21.64}{1} \\

MasaCtrl & SD 1.4
& \hval{0.028}{0.019}{0.079}{0}
& \hval{20.76}{16.90}{24.60}{1}
& \hval{0.105}{0.065}{0.228}{0}
& \hval{0.0112}{0.0083}{0.0278}{0}
& \hval{0.776}{0.610}{0.892}{1}
& \hval{22.85}{22.85}{25.74}{1}
& \hval{19.64}{19.58}{21.64}{1} \\

Pix2Pix-Zero & SD 1.4
& \hval{0.079}{0.019}{0.079}{0}
& \hval{16.90}{16.90}{24.60}{1}
& \hval{0.228}{0.065}{0.228}{0}
& \hval{0.0278}{0.0083}{0.0278}{0}
& \hval{0.677}{0.610}{0.892}{1}
& \hval{23.15}{22.85}{25.74}{1}
& \hval{19.58}{19.58}{21.64}{1} \\

PnP & SD 1.5
& \hval{0.025}{0.019}{0.079}{0}
& \hval{20.60}{16.90}{24.60}{1}
& \hval{0.109}{0.065}{0.228}{0}
& \hval{0.0107}{0.0083}{0.0278}{0}
& \hval{0.773}{0.610}{0.892}{1}
& \hval{24.33}{22.85}{25.74}{1}
& \hval{20.69}{19.58}{21.64}{1} \\

PnP-DirectInv & SD 1.5
& \hsecond{0.024}{0.019}{0.079}{0}
& \hval{20.82}{16.90}{24.60}{1}
& \hval{0.108}{0.065}{0.228}{0}
& \hsecond{0.0103}{0.0083}{0.0278}{0}
& \hval{0.774}{0.610}{0.892}{1}
& \hval{24.45}{22.85}{25.74}{1}
& \hval{20.69}{19.58}{21.64}{1} \\

LEDits++ & SD 1.5
& \hval{0.044}{0.019}{0.079}{0}
& \hval{18.94}{16.90}{24.60}{1}
& \hval{0.124}{0.065}{0.228}{0}
& \hval{0.0158}{0.0083}{0.0278}{0}
& \hval{0.750}{0.610}{0.892}{1}
& \hbest{25.74}{22.85}{25.74}{1}
& \hbest{21.64}{19.58}{21.64}{1} \\

ChordEdit & SD-Turbo
& \hval{0.030}{0.019}{0.079}{0}
& \hval{20.40}{16.90}{24.60}{1}
& \hval{0.139}{0.065}{0.228}{0}
& \hval{0.0120}{0.0083}{0.0278}{0}
& \hval{0.717}{0.610}{0.892}{1}
& \hval{24.00}{22.85}{25.74}{1}
& \hval{20.22}{19.58}{21.64}{1} \\

\addlinespace[0.12em]
\cmidrule(lr){1-9}

\rowcolor{datasetgray}
\multicolumn{9}{@{}l}{%
\hspace{0.25em}\textbf{\textit{Flow Methods}}} \\[-0.1em]

FlowEdit & FLUX.1-dev
& \hbest{0.019}{0.019}{0.079}{0}
& \hsecond{22.06}{16.90}{24.60}{1}
& \hsecond{0.092}{0.065}{0.228}{0}
& \hbest{0.0083}{0.0083}{0.0278}{0}
& \hsecond{0.843}{0.610}{0.892}{1}
& \hval{23.72}{22.85}{25.74}{1}
& \hval{20.05}{19.58}{21.64}{1} \\

RF-Inversion & FLUX.1-dev
& \hval{0.045}{0.019}{0.079}{0}
& \hval{18.07}{16.90}{24.60}{1}
& \hval{0.228}{0.065}{0.228}{0}
& \hval{0.0195}{0.0083}{0.0278}{0}
& \hval{0.610}{0.610}{0.892}{1}
& \hval{23.59}{22.85}{25.74}{1}
& \hval{20.06}{19.58}{21.64}{1} \\

\addlinespace[0.12em]
\cmidrule(lr){1-9}

\rowcolor{datasetgray}
\multicolumn{9}{@{}l}{%
\hspace{0.25em}\textbf{\textit{GRN Method}}} \\[-0.1em]

\textbf{\textit{RefineEdit (Ours)}} & GRN
& \hval{0.030}{0.019}{0.079}{0}
& \hbest{24.60}{16.90}{24.60}{1}
& \hbest{0.065}{0.065}{0.228}{0}
& \hval{0.0112}{0.0083}{0.0278}{0}
& \hbest{0.892}{0.610}{0.892}{1}
& \hsecond{25.12}{22.85}{25.74}{1}
& \hsecond{21.19}{19.58}{21.64}{1} \\

\bottomrule
\end{tabular}%
}
\end{table*}

\section{Algorithm of \textbf{\textit{RefineEdit}}}
\label{app:algorithm}

Algorithm~\ref{alg:refineedit} summarizes \textbf{\textit{RefineEdit}}
using the notation introduced in the main text.
$\operatorname{SampleBits}(P)$ samples each binary coordinate
from its predicted distribution over $\{0,1\}$.
Bernoulli sampling is applied independently to each coordinate.
The two branches share a fixed random code $Z$ and the same
refinement schedule, but use separate random streams.
For brevity, $G_\theta$ includes the standard GRN inference
settings for guidance and temperature.
The decoder $\mathcal{D}$ includes conversion from HBQ bits
to image features.

The algorithm performs $T$ prediction steps indexed from $0$ to $T-1$.
For $t<T-1$, $\lambda_{t+1}$ controls the prediction fraction
used to construct the next input.
At $t=T-1$, we update the editing mask and directly decode
the routed sampled predictions, without constructing another
input state or using $\lambda_T$.
In the algorithm, $\bar{a}_t$ denotes spatial selection before
freezing, and $\bar{m}_t$ denotes instantaneous bit selection
before locking.
The initial response $r$ is computed once at $t_s$ and remains
unchanged throughout subsequent steps.

\begin{algorithm}[htbp]
\footnotesize
\SetAlgoNlRelativeSize{0}
\caption{\textbf{\textit{RefineEdit}}: Training-Free Prompt-to-Prompt Image Editing}
\label{alg:refineedit}

\KwIn{
Frozen GRN $G_\theta$ and HBQ decoder $\mathcal{D}$;
source prompt $c^s$ and editing prompt $c^e$;
total prediction steps $T\geq 2$, switch step $0\leq t_s<T$,
and prediction-ratio schedule $\{\lambda_t\}_{t=0}^{T-1}$
with $\lambda_0=0$;
thresholds $\tau_{\mathrm{spa}}$, $\tau_{\mathrm{pow}}$,
and $\tau_{\mathrm{frz}}$;
bit-lock duration $K\geq 1$.
}
\KwOut{Source image $I^s$ and edited image $I^e$.}

$Z_{n,d}\sim\operatorname{Bernoulli}(1/2)$
for every binary coordinate $(n,d)$\;
$Y_0^s\gets Z$\;

\tcp{Source-only refinement}
\For{$t=0,\ldots,t_s-1$}{
    $P_t^s\gets
    \operatorname{softmax}\!\left(G_\theta(Y_t^s,c^s,t)\right)$
    \tcp*{Eq.~\ref{eq:grn_prediction}}

    $\widehat{Y}_t^s\gets
    \operatorname{SampleBits}(P_t^s)$\;

    $S_{t,n,d}^s\sim
    \operatorname{Bernoulli}(\lambda_{t+1})$\;

    $Y_{t+1}^s\gets
    S_t^s\odot\widehat{Y}_t^s+(1-S_t^s)\odot Z$
    \tcp*{Eq.~\ref{eq:grn_refinement}}
}

$Y_{t_s}^e\gets Y_{t_s}^s$
\tcp*{Shared-state branching; Sec.~\ref{sec:bit_routing}}

\tcp{Coupled edit localization and refinement}
\For{$t=t_s,\ldots,T-1$}{
    \For{$u\in\{s,e\}$}{
        $P_t^u\gets
        \operatorname{softmax}\!\left(G_\theta(Y_t^u,c^u,t)\right)$
        \tcp*{Eq.~\ref{eq:grn_prediction}}

        $\widehat{Y}_t^u\gets
        \operatorname{SampleBits}(P_t^u)$\;
    }

    \tcp{Compare probabilities of the same source-sampled bit}
    $\widehat{y}_{t,n,d}\gets\widehat{Y}_{t,n,d}^s$\;

    $\Delta_{t,n,d}\gets
    P_t^s(n,d,\widehat{y}_{t,n,d})
    -
    P_t^e(n,d,\widehat{y}_{t,n,d})$
    \tcp*{Eq.~\ref{eq:signed_probability_drop}}

    $q_{t,n}\gets
    \frac{1}{D}\sum_{d=1}^{D}\Delta_{t,n,d}$\;

    $\bar{a}_{t,n}\gets
    \mathbf{1}[q_{t,n}>\tau_{\mathrm{spa}}]$
    \tcp*{Eq.~\ref{eq:spatial_bit_masks}}

    \tcp{AdaSF: compute the initial response only once}
    \If{$t=t_s$}{
        $\Omega\gets\{n\mid\bar{a}_{t_s,n}=1\}$\;

        $r\gets
        \frac{
            \sum_{n\in\Omega}(q_{t_s,n}-\tau_{\mathrm{spa}})
        }{
            \max(1,|\Omega|)
        }$
        \tcp*{Eq.~\ref{eq:initial_response}}
    }

    \tcp{Apply the fixed freezing decision at every editing step}
    \eIf{$r\geq\tau_{\mathrm{frz}}$}{
        $a_t\gets\bar{a}_{t_s}$
        \tcp*{Reuse the initial spatial mask}
    }{
        $a_t\gets\bar{a}_t$
        \tcp*{Use the current spatial mask}
    }

    $\bar{m}_{t,n,d}\gets
    a_{t,n}\,\mathbf{1}[\Delta_{t,n,d}>\tau_{\mathrm{pow}}]$
    \tcp*{Eq.~\ref{eq:spatial_bit_masks}}

    \tcp{FBL: retain recent editing permissions}
    $m_t\gets
    \bigvee_{j=\max(t_s,t-K+1)}^{t}\bar{m}_j$
    \tcp*{Eq.~\ref{eq:finite_bit_lock}}

    \tcp{Construct the next input only before the final step}
    \If{$t<T-1$}{
        \For{$u\in\{s,e\}$}{
            $S_{t,n,d}^u\sim
            \operatorname{Bernoulli}(\lambda_{t+1})$\;

            $\widetilde{Y}_{t+1}^u\gets
            S_t^u\odot\widehat{Y}_t^u
            +(1-S_t^u)\odot Z$
            \tcp*{Eq.~\ref{eq:grn_refinement}}
        }

        $Y_{t+1}^s\gets\widetilde{Y}_{t+1}^s$\;

        $Y_{t+1}^e\gets
        m_t\odot\widetilde{Y}_{t+1}^e
        +(1-m_t)\odot Y_{t+1}^s$
        \tcp*{Eq.~\ref{eq:source_anchored_routing}}
    }
}

\tcp{Decode final sampled predictions without mixing with $Z$}
$I^s\gets\mathcal{D}(\widehat{Y}_{T-1}^s)$\;

$I^e\gets
\mathcal{D}\!\left(
m_{T-1}\odot\widehat{Y}_{T-1}^e
+(1-m_{T-1})\odot\widehat{Y}_{T-1}^s
\right)$\;

\Return{$I^s,I^e$}\;

\end{algorithm}

\clearpage
\section{Additional Qualitative Results}
\label{app:qualitative_results}

\begin{figure}[!htbp]
    \centering
    \includegraphics[
        width=\linewidth,
        height=0.75\textheight,
        keepaspectratio
    ]{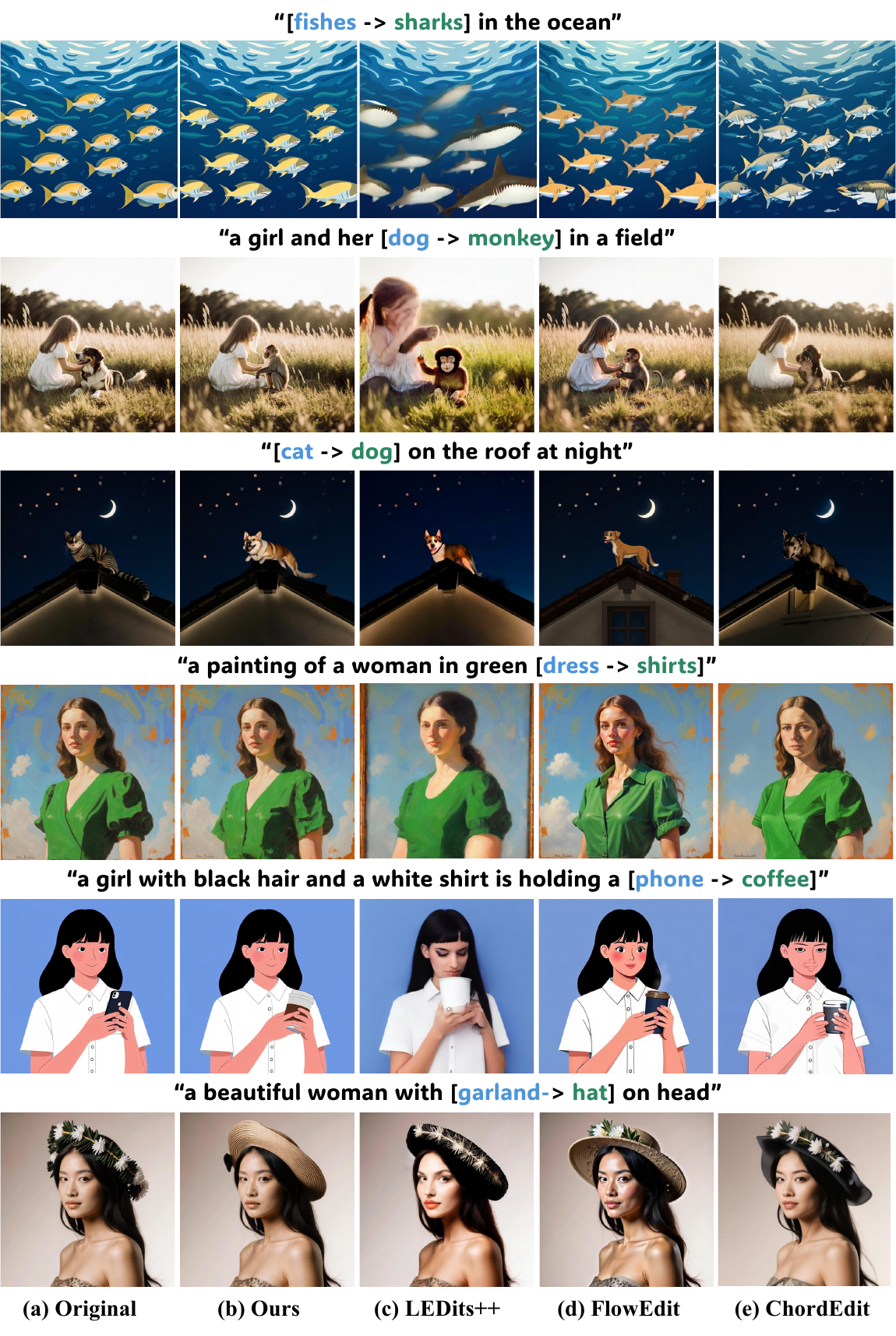}
    \caption{
    Qualitative comparison on object replacement.
    The editing prompt replaces a source object with another category
    while leaving unrelated scene content unchanged.
    }
    \label{fig:app_performance01}
\end{figure}

\begin{figure}[!htbp]
    \centering
    \includegraphics[
        width=\linewidth,
        height=0.82\textheight,
        keepaspectratio
    ]{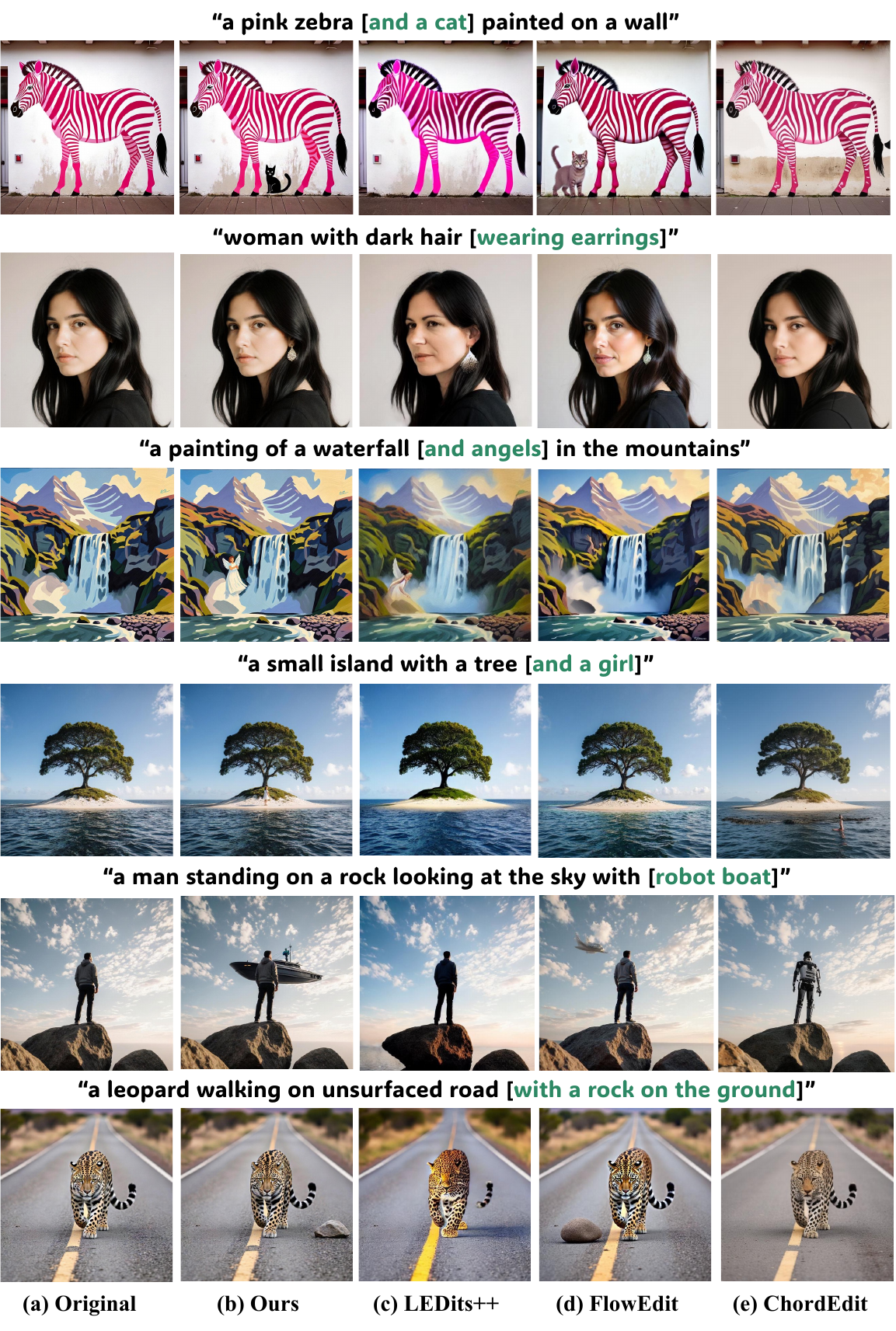}
    \caption{
    Qualitative comparison on object addition.
    The task introduces new objects specified by the editing prompt
    while preserving existing content and integrating the additions
    into the scene.
    }
    \label{fig:app_performance02}
\end{figure}

\begin{figure}[!htbp]
    \centering
    \includegraphics[
        width=\linewidth,
        height=0.82\textheight,
        keepaspectratio
    ]{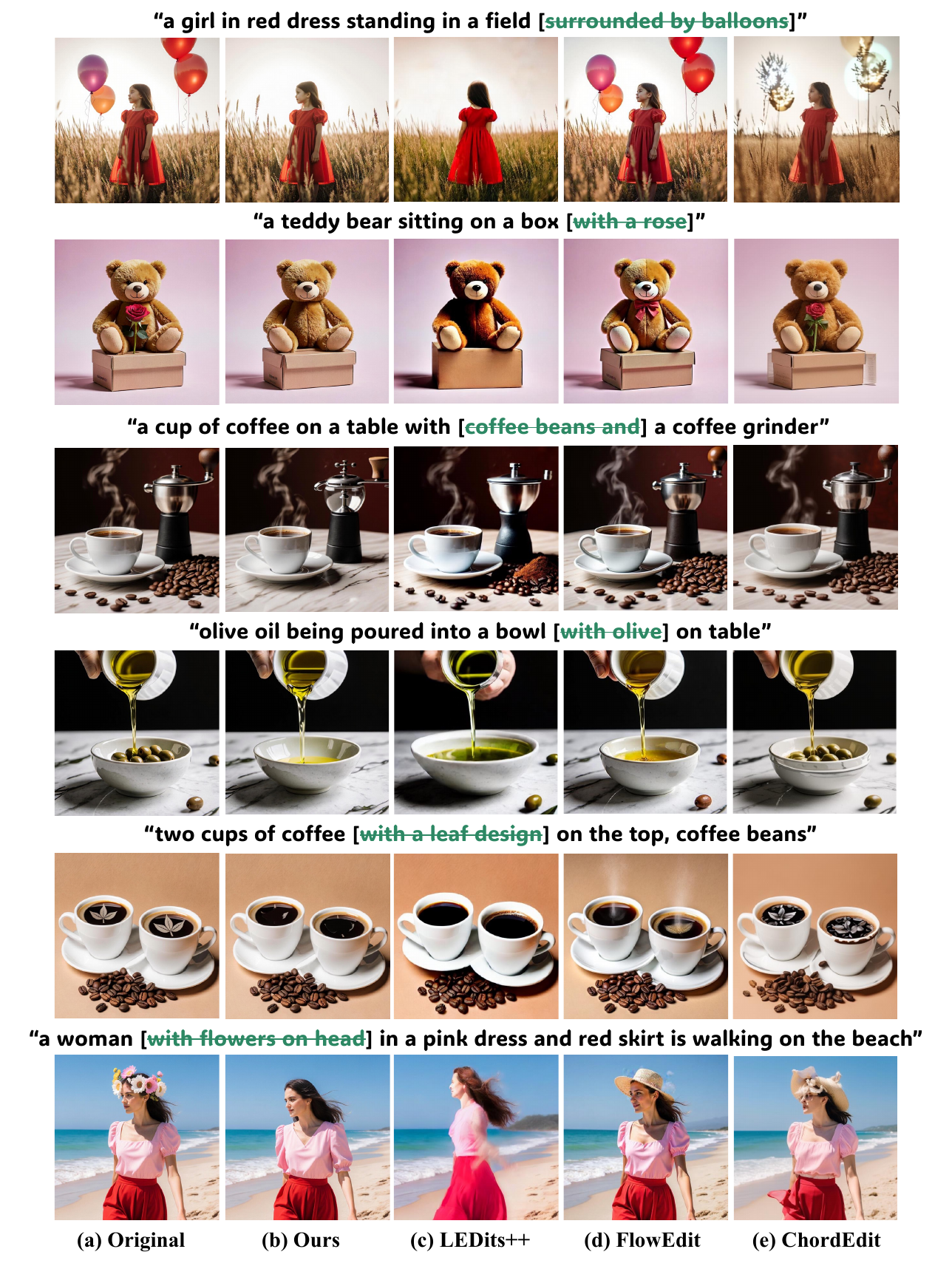}
    \caption{
    Qualitative comparison on object removal.
    The specified object should be removed and its former region
    filled coherently, without unnecessary changes elsewhere.
    }
    \label{fig:app_performance03}
\end{figure}

\begin{figure}[!htbp]
    \centering
    \includegraphics[
        width=\linewidth,
        height=0.82\textheight,
        keepaspectratio
    ]{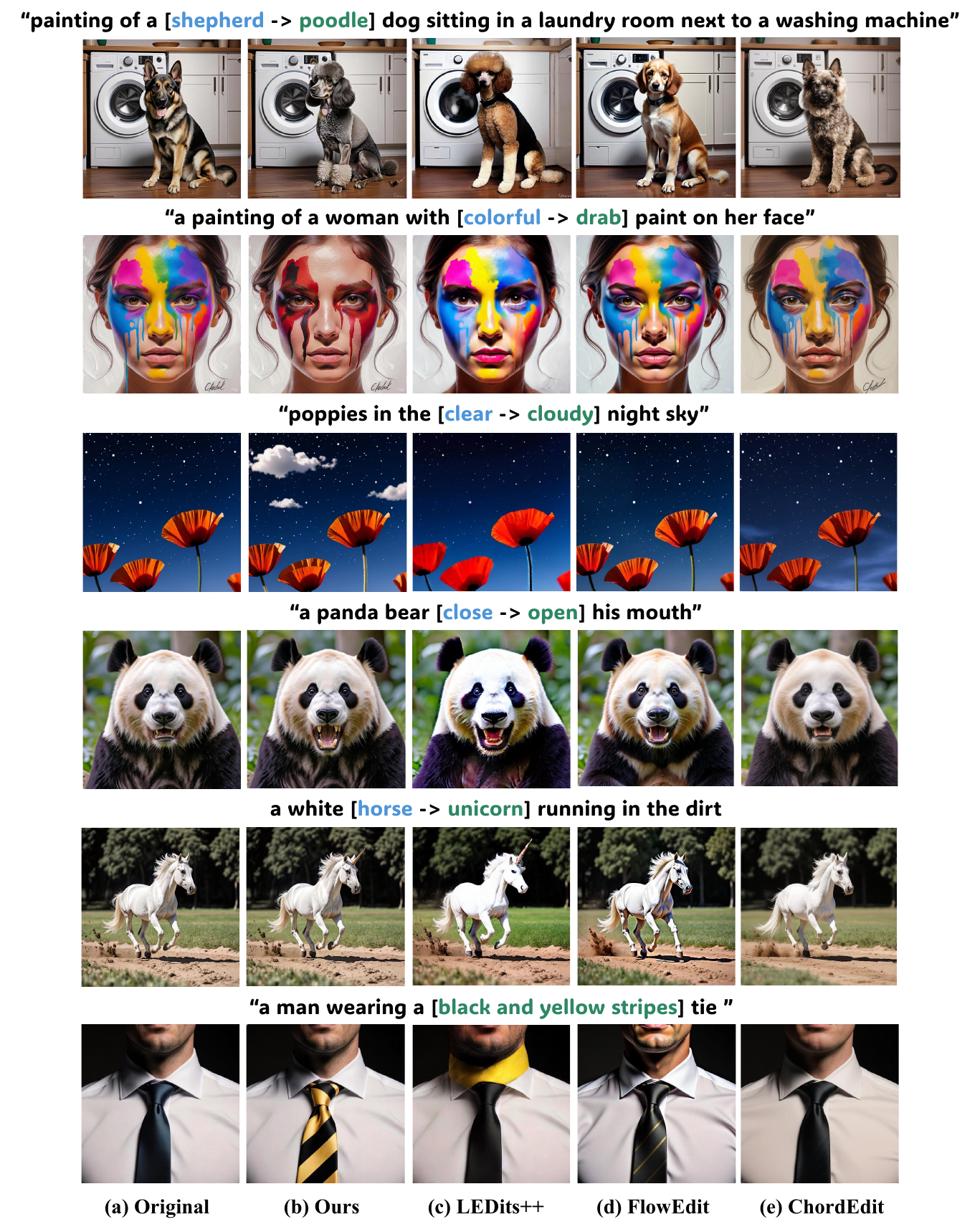}
    \caption{
    Qualitative comparison on content modification.
    The task changes an object's attributes or internal appearance,
    such as its expression or pattern, while retaining unrelated
    object and scene characteristics.
    }
    \label{fig:app_performance04}
\end{figure}

\begin{figure}[!htbp]
    \centering
    \includegraphics[
        width=\linewidth,
        height=0.82\textheight,
        keepaspectratio
    ]{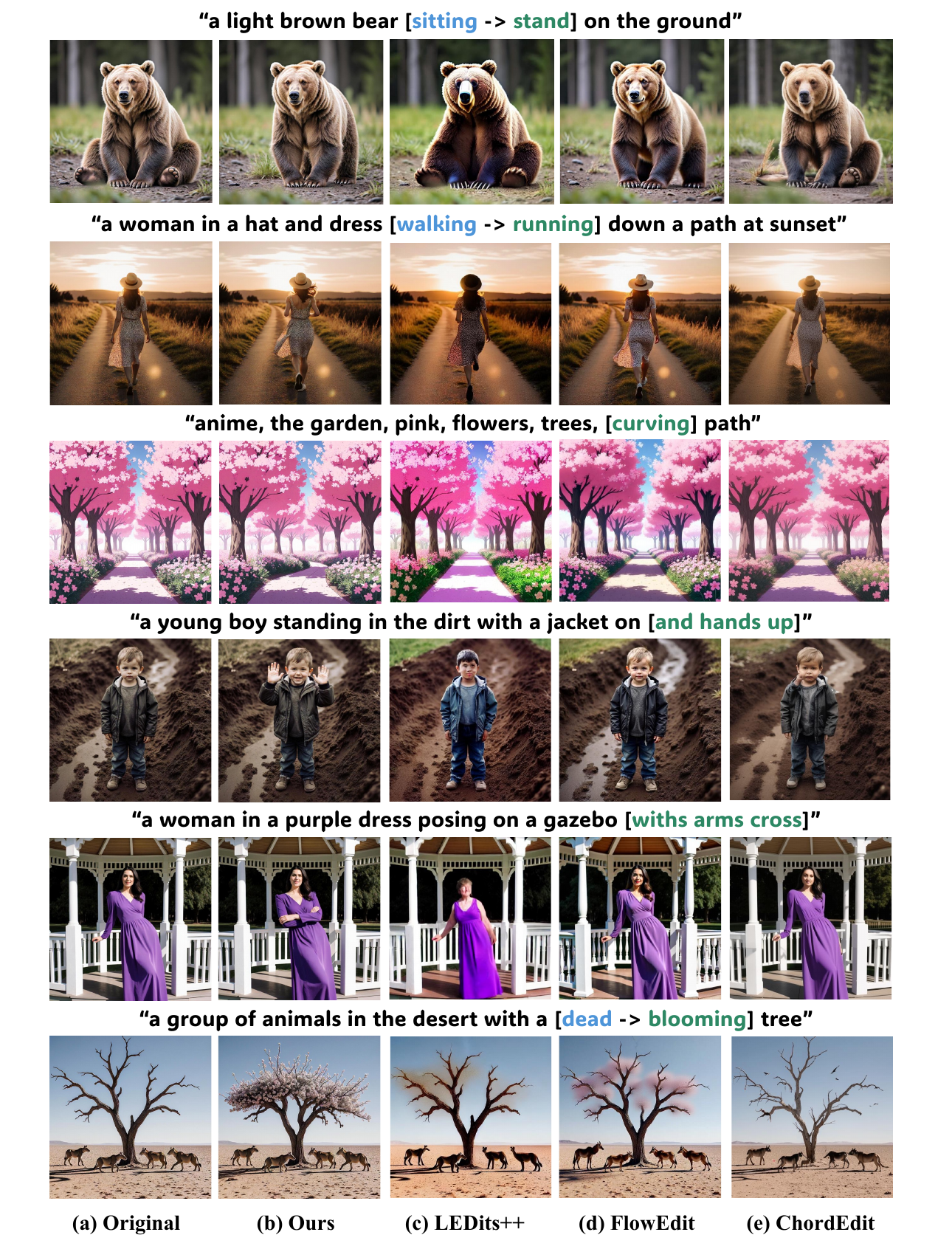}
    \caption{
    Qualitative comparison on pose modification.
    The editing prompt changes a subject's pose, action, or orientation.
    Successful editing requires the corresponding structural changes
    while maintaining a coherent subject and scene.
    }
    \label{fig:app_performance05}
\end{figure}

\begin{figure}[!htbp]
    \centering
    \includegraphics[
        width=\linewidth,
        height=0.82\textheight,
        keepaspectratio
    ]{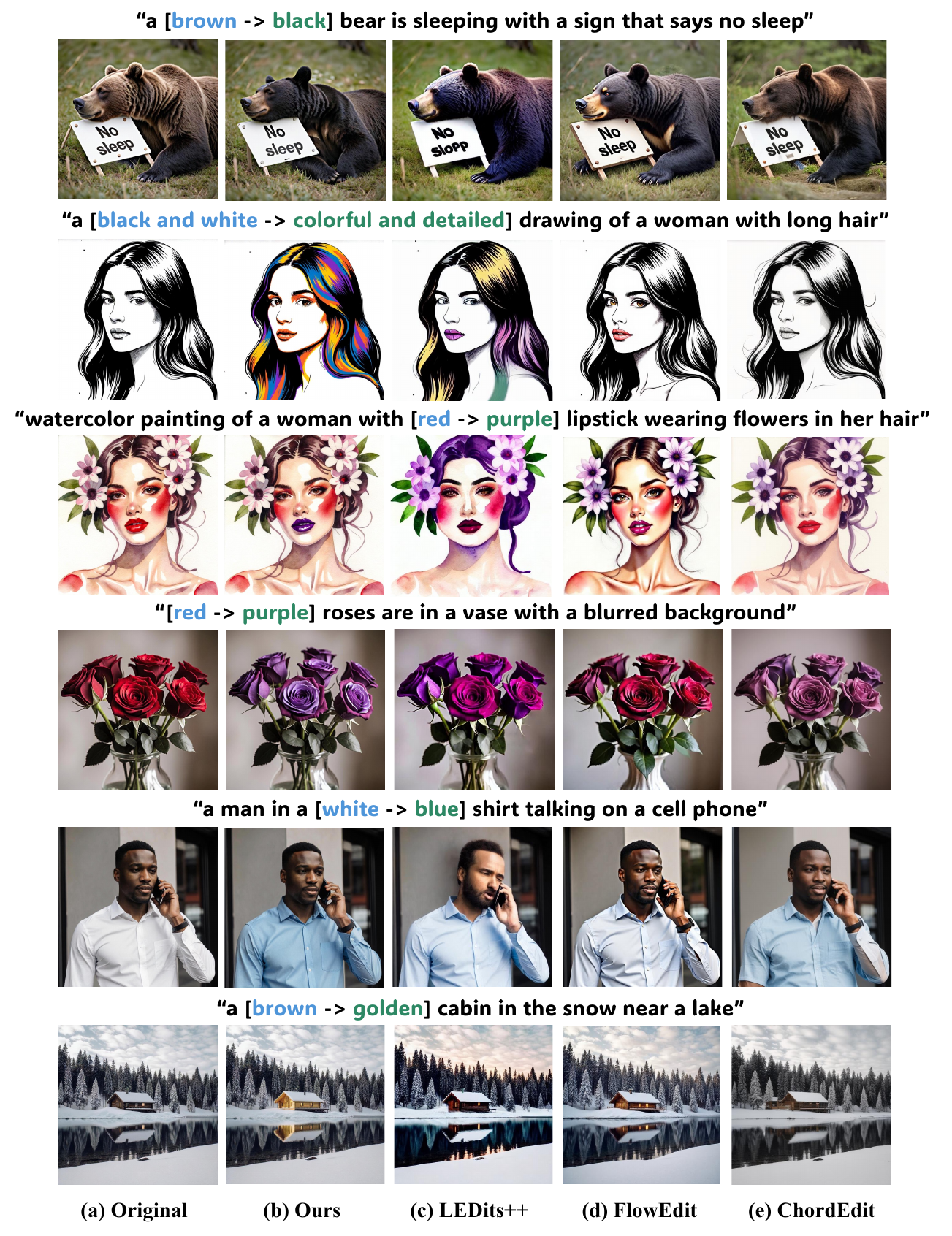}
    \caption{
    Qualitative comparison on color modification.
    The specified color change should remain localized to the target
    object while preserving its shape and the appearance of
    unrelated regions.
    }
    \label{fig:app_performance06}
\end{figure}

\begin{figure}[!htbp]
    \centering
    \includegraphics[
        width=\linewidth,
        height=0.82\textheight,
        keepaspectratio
    ]{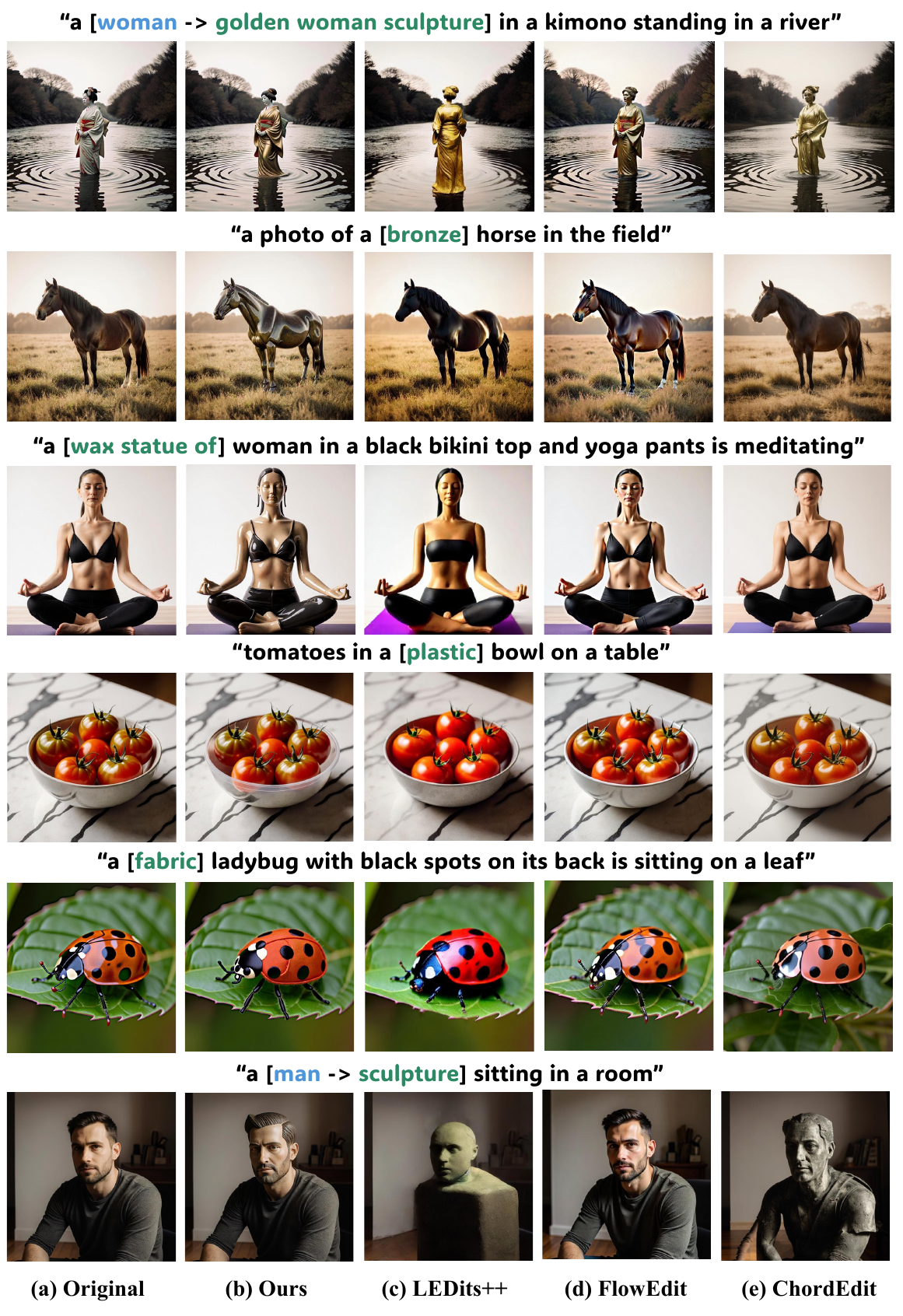}
    \caption{
    Qualitative comparison on material modification.
    The task changes the target object's material appearance,
    including its texture and reflectance, while retaining its
    overall form and surrounding content.
    }
    \label{fig:app_performance07}
\end{figure}

\begin{figure}[!htbp]
    \centering
    \includegraphics[
        width=\linewidth,
        height=0.82\textheight,
        keepaspectratio
    ]{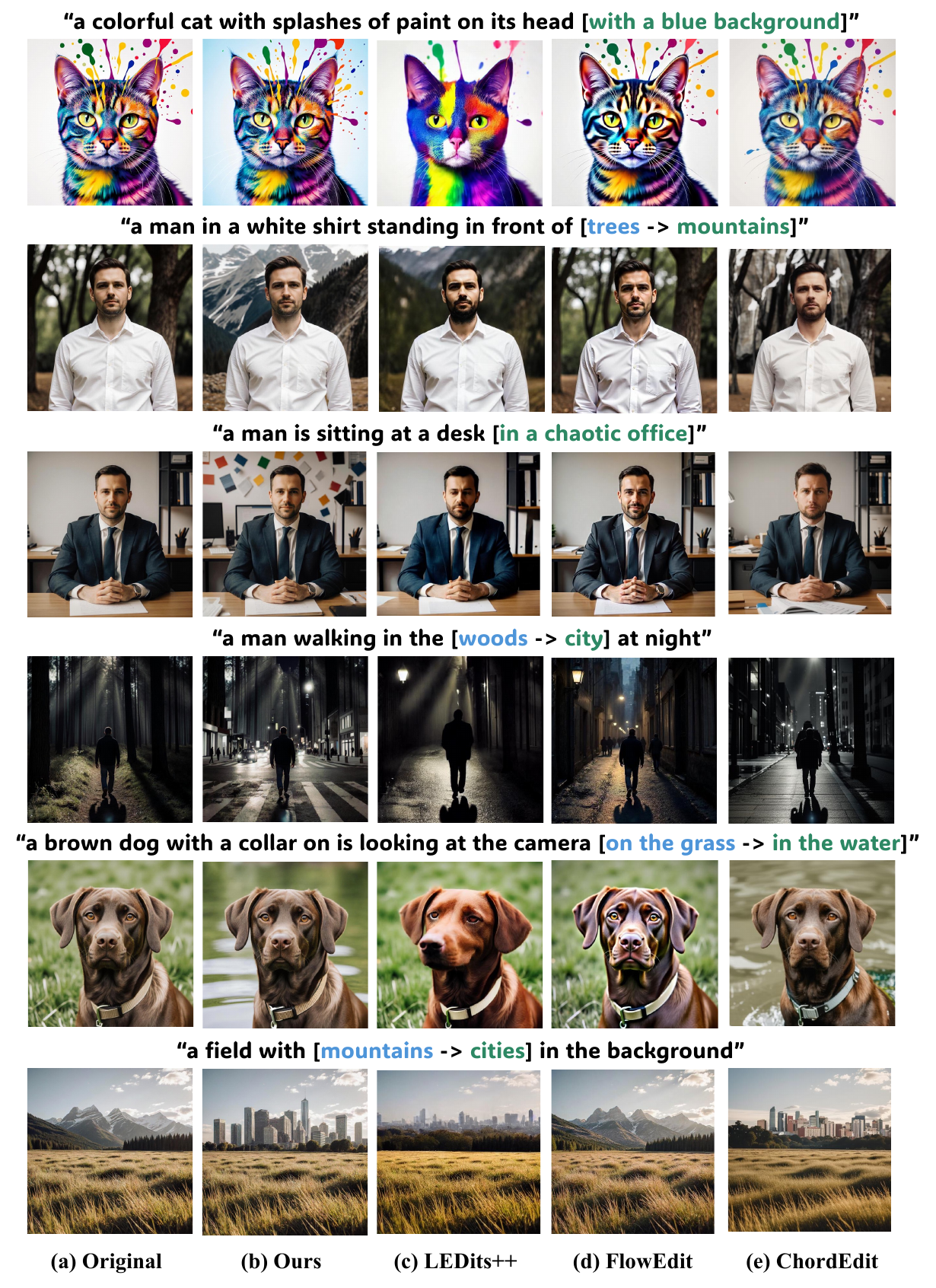}
    \caption{
    Qualitative comparison on background modification.
    The surrounding scene is changed according to the editing prompt,
    while the foreground subject should remain consistent with
    the source image.
    }
    \label{fig:app_performance08}
\end{figure}

\begin{figure}[!htbp]
    \centering
    \includegraphics[
        width=\linewidth,
        height=0.82\textheight,
        keepaspectratio
    ]{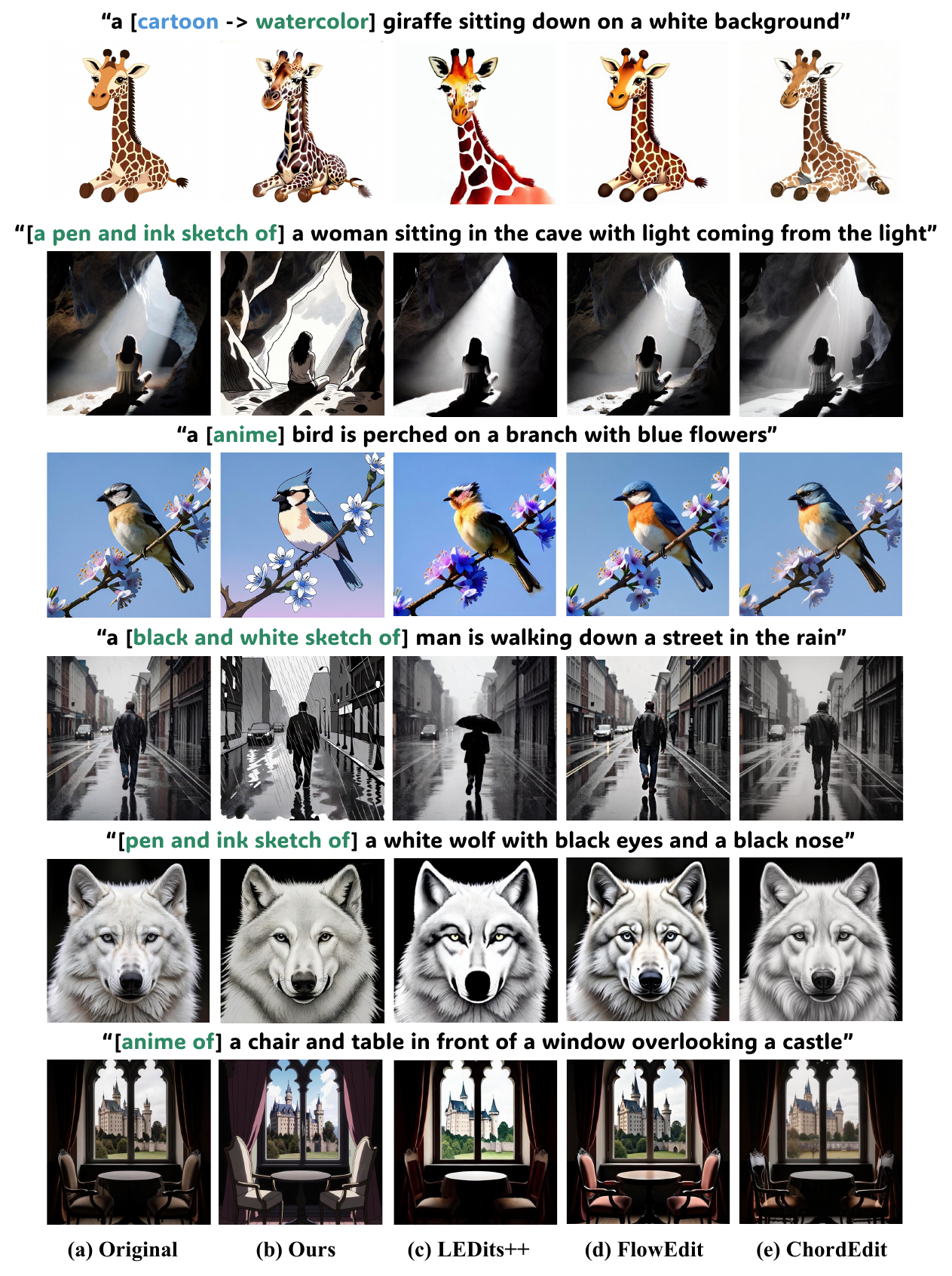}
    \caption{
    Qualitative comparison on style transfer.
    The target style is applied across the image while preserving
    recognizable scene content and composition.
    Unlike localized editing, this task permits changes throughout
    the image.
    }
    \label{fig:app_performance09}
\end{figure}

We provide additional qualitative comparisons across the nine
editing categories in our evaluation set.
Figures~\ref{fig:app_performance01}--\ref{fig:app_performance09}
show how different methods introduce the requested changes
while preserving content unrelated to the editing prompt.
These examples complement the quantitative evaluation in the main text.


\clearpage

\clearpage

\clearpage

\clearpage

\clearpage

\clearpage

\clearpage

\clearpage

\clearpage

\end{document}